\documentclass[10pt]{article} 
\usepackage[preprint]{tmlr}

\usepackage{hyperref}
\usepackage{url}

\usepackage[T1]{fontenc}
\usepackage[utf8]{inputenc}
\usepackage{microtype}
\usepackage{inconsolata}

\usepackage[most]{tcolorbox}
\usepackage{listings}
\usepackage{graphicx}
\usepackage{booktabs}
\usepackage{amsmath}
\usepackage{amssymb}
\usepackage{enumitem}
\usepackage{subcaption}
\usepackage{multirow}
\usepackage{xcolor}

\title{Are Finer Citations Always Better? \\ Rethinking Granularity for Attributed Generation}

\author{\name Hexuan Wang \email hwang302@jhu.edu \\
      \name Jingyu Zhang\thanks{Equal advising.} \email jzhan237@jhu.edu \\
      \name Benjamin Van Durme \email vandurme@jhu.edu \\
      \name Daniel Khashabi\footnotemark[1] \email danielk@jhu.edu \\
      \addr Department of Computer Science, Johns Hopkins University}

\def\month{MM}  
\def\year{YYYY} 
\def\openreview{\url{https://openreview.net/forum?id=XXXX}} 

\begin{document}

\maketitle

\begin{abstract}
Citation granularity—whether to cite individual sentences, paragraphs, or documents—is a critical design choice in attributed generation. While fine-grained citations are commonly preferred for precise human verification, their impact on model performance remains under-explored. We analyze four model scales (8B–120B) and demonstrate that enforcing fine-grained (sentence-level) citations forfeits gains of 2–97\% in attribution quality (median 40\%) relative to the best-performing granularity, and up to 338\% on individual tasks. Strikingly, setting citation granularity to its optimal value (based on attribution quality) unlocks these substantial gains while leaving overall answer correctness essentially unchanged (between $-2.3\%$ and $+4.4\%$). We observe a consistent pattern where attribution quality peaks at intermediate (paragraph-level) granularities: finer citations appear to sever the semantic dependencies needed to ground a claim, while excessively coarse citations introduce distracting noise. Importantly, this performance gap varies with scale: when a claim rests on a small or moderate amount of evidence, it disproportionately penalizes larger models by disrupting the multi-sentence information synthesis at which they excel. Fine-grained citation rests on the premise that a sentence is a sufficient unit of evidence on its own. Our results indicate that it often is not, and that this is a property of the model rather than of the citation standard. Standards fixed for human verifiability may therefore paradoxically degrade the very attribution they aim to ensure; effective attribution requires matching granularity to the model's semantic scope rather than fixing it by convention.
\end{abstract}

\section{Introduction}
\begin{figure}[t]
    \centering
    \includegraphics[width=\textwidth]{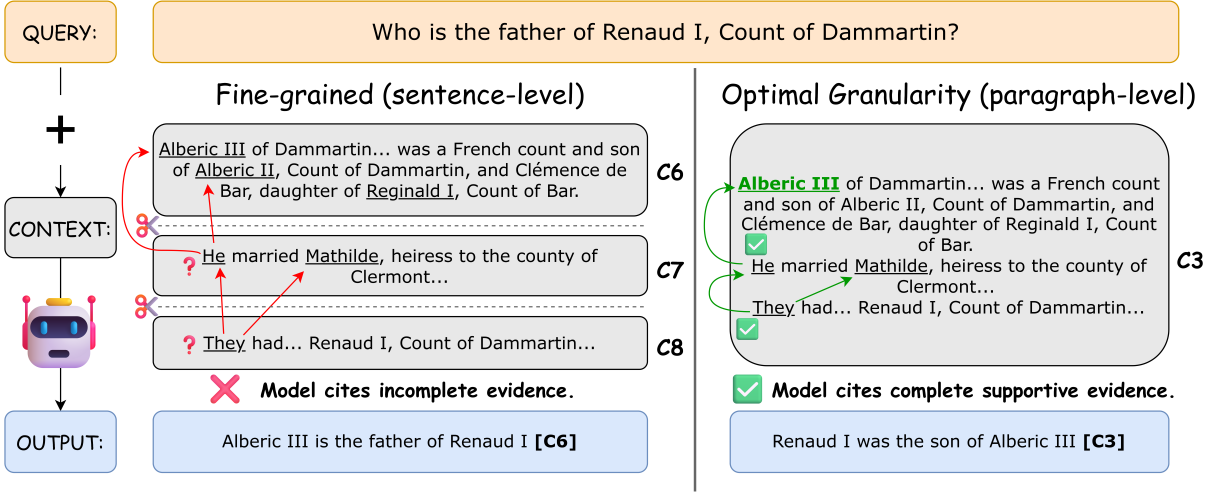} 
    \caption{\textbf{Impact of Citation Granularity.} The model processes a long context with many chunks and generates an answer with supporting citations. We show only the chunks relevant to the query. At fine granularity (Left), sentence-level chunks (C6-C8) cannot independently support the statement. Although the generated answer is correct, attribution fails because fine-grained chunking isolates the subject (``Alberic III'') from the relationship. The model selects the citation chunk containing the subject anchor (C6), but this evidence is incomplete without the dependency chain. In contrast, coarse granularity (Right) merges these sentences into paragraph-level chunk C3, preserving the full dependency required for a supportive citation.}
    \label{fig:granularity_mismatch}
\end{figure}
Grounding Large Language Model (LLM) outputs in verifiable sources is an important mechanism for establishing trustworthiness~\citep{maynez-etal-2020-faithfulness, ji2022survey}. Attributed generation systems typically require models to cite specific text segments of retrieved documents~\citep{nakano2021webgpt,gao2023enabling}. A common design choice is fine-grained citation---often individual sentences or short text spans~\citep{menick2022teaching,bohnet2022attributed,zhang2024longcite,berchansky-etal-2024-cotar}. The rationale is that fine-grained citations facilitate human verification by pinpointing claims to specific evidence supporting them~\citep{cao-wang-2024-verifiable,chen2025concisesufficientsubsentencecitations}. Consequently, benchmarks like ALCE~\citep{gao2023enabling} and LongBench-Cite~\citep{zhang2024longcite} have adopted \textit{fine-grained} citation as standard.

However, this design choice implicitly prioritizes human verification—the simplicity of matching generated statements to sources—over the model's capacity for information synthesis. Natural language is rarely semantically atomic; understanding claims often requires context spanning multiple sentences~\citep{weng-etal-2023-large, arafat2025citationgroundedcodecomprehensionpreventing}. We hypothesize that treating individual sentences as atomic units introduces challenges beyond simple retrieval: such constraints disrupt the coreference chains and multi-sentence reasoning structures required for faithful generation. Our findings are consistent with this hypothesis: the fine-grained constraint introduced to facilitate human verification imposes a severe performance burden on the very attribution it was designed to ensure (Figure~\ref{fig:granularity_mismatch}).

Investigating this hypothesis requires addressing a critical confound: citation granularity is inherently tied to \textbf{citation volume} (the total number of sentences cited). 
Without controlling for volume, coarser-grained citations would naturally encompass more total text than fine-grained ones, creating an imbalance in the specific evidence available for grounding and verification. 
To isolate the effect of granularity, we control for citation volume throughout. By comparing models only when they cite equivalent amounts of text, we ensure that performance differences stem specifically from \textbf{how the evidence is segmented} rather than \textbf{how much evidence is provided}. Our analysis across four model scales (8B--120B) reveals several key findings.

First, we observe a consistent pattern: \textbf{attribution quality (measured by Citation F1) is lowest at sentence-level, peaks at paragraph-level, and stabilizes or declines at multi-paragraph granularity}. Across all four models and matched citation volumes, enforcing fine-grained citation forfeits gains of 2--97\% in attribution quality (median 40\%) compared to the best-performing granularity. We identify two competing failure modes driving this pattern (\S\ref{sec:mechanism}).
At fine granularities, we hypothesize that isolating sentences into atomic units fractures the semantic context required for synthesis, a phenomenon aligning with known dependencies in complex reasoning~\citep{weng-etal-2023-large, arafat2025citationgroundedcodecomprehensionpreventing}. Conversely, at coarse granularities, excessively large segments introduce irrelevant information, a phenomenon we suggest is analogous to the performance degradation observed in long-context settings~\citep{liu2023lost,pmlr-v202-shi23a}.

Second, and most strikingly, \textbf{these attribution gains come at essentially no cost to answer correctness.} Across all models, citation-optimal granularity yields massive attribution gains (+18--59\%) with negligible impact on overall correctness ($-2.3\%$ to $+4.4\%$). This asymmetry indicates that while attribution is highly sensitive to granularity, the model's core reasoning remains stable. For capable models (e.g., Llama-70B), citation-optimal granularity even improves correctness across all volume regimes (+0.7\% to +9.2\%), though this relationship varies by model and volume. These findings refute concerns that coarser citation encourages imprecise or ``lazy'' responses; instead, they reveal that fine-grained standards impose severe attribution penalties for negligible correctness benefits.

Third, \textbf{the magnitude of this performance gap varies with model scale, disproportionately penalizing larger models.} In the medium-volume regime ($16 \le V \le 31$), where this pattern is most pronounced, Llama-8B gains only +2.2\% from the best-performing granularity while Llama-70B gains +70.2\%; in the GPT-OSS family, the 120B model gains +96.7\% versus +34.0\% for the 20B model. We hypothesize that fine-grained constraints disproportionately suppress the superior synthesis capabilities of large models, preventing them from leveraging the broad contexts required for complex reasoning. A practical consequence is that the granularity a benchmark fixes can reorder the models it ranks: at $k{=}2$ Llama-8B appears to attribute more reliably than both Llama-70B and GPT-120B (0.536 against 0.398 and 0.393), an ordering that both larger models reverse at $k{=}4$.

{Ultimately, these findings expose a critical flaw in current attribution design: fine-grained standards, enforced for human verifiability, paradoxically degrade the attribution they aim to ensure. We provide the first systematic quantitative evidence of this misalignment across four model scales. This work motivates a new evaluation paradigm: citation granularity must be treated as a dynamic, model-centric parameter to be empirically measured, not a rigid standard prescribed by benchmarks.}

\section{Related Work}


\paragraph{Attribution and Citation Benchmarks.}
\citet{gao2023enabling} introduced ALCE, the first large-scale benchmark for attributed text generation, establishing short-text-span citation as the evaluation standard. \citet{zhang2024longcite} extended this paradigm to long-context settings with LongBench-Cite, maintaining fine-grained citation. 
\citet{bohnet2022attributed} developed attributed question answering, similarly adopting sentence-level granularity. 
While recent studies such as SciFi~\citep{cao-wang-2024-verifiable} and ALiiCE~\citep{xu-etal-2025-aliice} have also explored citation granularity, they focus strictly on \textit{positional} granularity (e.g., evaluating whether placing citation markers mid-sentence or at the end of a claim best facilitates human verification), rather than the length of the cited text segments themselves.
Furthermore, a survey by \citet{schreieder2025attributioncitationquotationsurvey} highlights a growing trend in the literature toward ``atomic'' verifiability, where sentence and token-level constraints are increasingly prioritized to facilitate precise human review. Our work is the first to systematically question this assumption by controlling for citation volume, revealing that fine-grained constraints can degrade performance, particularly for capable models.


\paragraph{Long-Context Processing and Model Capabilities.}
Recent work examines how models process long contexts~\citep{liu2025comprehensive,wu2024retrievalheadmechanisticallyexplains}. \citet{liu2023lost} showed models struggle when relevant information is buried in long contexts (``Lost in the Middle''), while \citet{pmlr-v202-shi23a} demonstrated that irrelevant context can distract models. \citet{weng-etal-2023-large} studied how context dependencies affect reasoning. \citet{li2024longcontextllmsstrugglelong} found that even models with extended context windows struggle with long in-context learning. \citet{chen2025surveyscalinglargelanguage,wang2025largerlanguagemodelsgeneralize,wei2022emergent} demonstrated that model capabilities scale with size, with larger models showing emergent abilities. Our findings complement this literature by showing that as models develop stronger semantic synthesis capabilities, they exhibit greater performance variation across granularities in volume-dependent patterns, with fine-grained fragmentation disproportionately harming capable models in typical usage scenarios.

\paragraph{Retrieval Chunking and Semantic Segmentation.}
Work on retrieval-augmented generation explores optimal chunking strategies. \citet{zhong-etal-2025-mix} proposed Mix-of-Granularity, dynamically determines optimal chunk sizes for retrieval. While retrieval systems often use fine-grained context segmentation to improve search, recent findings show this actually \textbf{degrades retrieval accuracy}~\citep{jiang2024longragenhancingretrievalaugmentedgeneration,bianchi2025hiddenhaystacksmallerneedles, cheng2026sustainable}. Our work extends this insight to the \textit{generation} stage: we demonstrate that even when retrieval is perfect, fine-grained constraints fracture the semantic dependencies required for faithful generation and attribution.

\paragraph{Citation Generation Methods.}
Recent methods attempt to improve citation quality and verifiability through various training and generation strategies. \citet{ye-etal-2024-effective} proposed AGREE for dynamic evidence retrieval during generation. \citet{li-etal-2024-improving-attributed} demonstrated that preference learning can improve citation precision. Alternatively, \citet{zhang2025verifiable} propose Quote-Tuning to bypass retrieval entirely, aligning models to quote verbatim from pre-training data to enhance trustworthiness. However, for systems that do rely on retrieved contexts, existing methods implicitly assume finer granularity improves citation quality and focus on training techniques to achieve it. In contrast, we demonstrate that this assumption is incorrect for capable models, and that architectural choices around citation granularity fundamentally affect what models can achieve regardless of training method.

\section{Methodology}
\label{sec:method}

We introduce a formal framework to isolate citation granularity from confounding factors. We formalize the governing variables in \S\ref{subsec:variable}, address prompt bias and retrieval recall via a unified generation protocol in \S\ref{subsec:prompt_control}, and describe our stratification strategy for strictly controlling citation volume in \S\ref{subsec:volume-controlled}.  


\subsection{Problem Formalization}
\label{subsec:variable}

To systematically analyze this problem, we formalize the attribution process using two core variables: citation granularity ($k$) and citation volume ($V$). 

\paragraph{Citation Granularity ($k$).}
We define granularity $k$ as the size of evidence segments provided to the model, measured in sentences. The source document is pre-segmented into a sequence of chunks $C = \{c_0, c_1, \dots\}$, where each chunk $c_i$ contains exactly $k$ consecutive sentences; the model then attributes a generated statement $S$ by referencing a set of chunk indices $I_S$. As illustrated in Figure~\ref{fig:granularity_mismatch}, a sentence-level granularity (Left) chunks the document by individual sentences, isolating chunks such as C6, C7, and C8.  
Whereas a paragraph-level granularity (Right) groups these sentences into a single, cohesive unit, labeled as chunk C3. 

\paragraph{Citation Volume ($V$).}
We define volume $V$ as the total number of sentences cited for a statement $S$, calculated as the number of cited chunks multiplied by the granularity:
\(
    V(S) = |I_S| \times k.
\)
For example, in Figure~\ref{fig:granularity_mismatch}, citing three distinct sentences (Left) and citing the single corresponding paragraph chunk (Right) both result in a volume of $V{=}3$, despite the difference in chunking.

\paragraph{The Confound.}
\textit{Granularity and volume are naturally confounded.} A model citing two chunks at $k{=}8$ accesses $V{=}16$ sentences, whereas at $k{=}1$ it accesses only $V{=}2$. To isolate the effect of how evidence is segmented ($k$) from the total evidence provided ($V$), we must control for $V$. We describe our strategy for strictly controlling this confound via post-hoc stratification in \S\ref{subsec:volume-controlled}.

\subsection{Task and Controlled Generation Protocol}
\label{subsec:prompt_control}
We adopt the standard joint attributed generation setting from LongBench-Cite~\citep{zhang2024longcite}, where the model receives a query and the \textit{complete} source document formatted as the chunk sequence $C$ (defined in \S\ref{subsec:variable}), generating a response with interleaved citations in a single inference pass.

Two properties of this protocol define the scope of our findings, and both are deliberate. First, generation and citation occur jointly rather than as separate stages. This is the protocol used by the benchmarks we critique and by practical attribution systems~\citep{nakano2021webgpt,menick2022teaching,zhang2024longcite}, so evaluating under it measures the paradigm actually in use; the cost is that our metrics do not separate reasoning difficulty from citation difficulty. Second, the model always receives the complete gold document. Introducing retrieval noise would conflate granularity effects with missing-evidence effects~\citep{chen2026does}, so we first isolate the generation stage. Our results therefore characterize an upper-bound regime in which the relevant evidence is guaranteed to be present. We return to both choices in \S\ref{sec:limitations}.

To rigorously study granularity effects, we employ a \textbf{Unified Granularity Prompting} strategy. We generalize the original in-context prompting—which utilizes fixed sentence-level ($k{=}1$) 
demonstrations—to a flexible format that supports variable granularity, thereby minimizing prompt-induced bias. We provide the complete source document formatted as sequence $C$ in the context window for all settings, manipulating only the granularity parameter $k \in \{1, 2, 4, 8, 16, 32\}$. 
Here, $k$ strictly dictates how the document is \textit{chunked} in the prompt (e.g., whether each chunk covers 1 sentence or 16), not what information is visible.

This design eliminates retrieval recall as a confound—the semantic content available to the model is identical regardless of $k$. Unlike retrieval-augmented systems where chunking affects  
which information is retrieved~\citep{zhong-etal-2025-mix}, our setup isolates the specific impact of citation granularity on the model's reasoning and attribution mechanics. The full generation prompt template is provided in Table~\ref{tab:prompt_generation} (Appendix~\ref{app:prompt}).

\subsection{Volume-Controlled Stratification}
\label{subsec:volume-controlled}
To address the volume confound defined in \S\ref{subsec:variable}, we employ \textbf{Volume-Controlled Stratification}. We stratify generated statements post-hoc based on the total sentences cited ($V$) into discrete ranges spanning 0, 1, 2--3, 4--7, 8--15, 16--31, 32--63, and 64+ sentences. We exclude $V{=}0$ (statements the model left uncited, which have no citation granularity), $V{=}1$, and $V{\ge}64$ from the primary analysis, due to insufficient granularity overlap or low sample sizes; the retained ranges cover 47,297 of the 60,976 generated statements.
Comparisons between granularity settings are conducted \textit{strictly within each volume range}, ensuring that models are compared only when supporting claims with equivalent amounts of text. 
We verify the validity of this control by confirming that mean citation volumes remain tightly matched across granularity settings within each volume range. For Llama-70B at $16 \le V \le 31$, for instance, the mean number of cited sentences is 19.1, 19.0, and 18.2 at $k{=}2$, $4$, and $8$ (between-dataset SD $\le 0.91$): statements compared across granularities cite near-identical amounts of text. Appendix~\ref{app:methodology} provides a detailed demonstration of the volume distributions and structural constraints such as the impossibility of citing a single chunk $k$ larger than the total volume $V$.


\section{Experimental Setup}\label{sec:setup}

\subsection{Models and Data}
We evaluate four instruction-tuned models spanning a 15$\times$ scale range: Llama-3.1-8B-Instruct~\citep{dubey2024llama3herdmodels} (8B), GPT-OSS-20B~\citep{openai2025gptoss120bgptoss20bmodel} (20B), Llama-3.3-70B-Instruct~\citep{dubey2024llama3herdmodels} (70B), and GPT-OSS-120B~\citep{openai2025gptoss120bgptoss20bmodel} (120B).

We evaluate on the English subset of \textbf{LongBench-Cite}~\citep{zhang2024longcite} (585 queries). To ensure robust domain generalization, we leverage its design as a meta-benchmark aggregating \textbf{four distinct datasets} that cover the primary paradigms of long-context generation. These feature extreme context lengths (dataset means from 4.6K to 17.5K words, with individual instances reaching 63K) spanning diverse domains: HotpotQA~\citep{yang2018hotpotqa} (multi-document Wikipedia reasoning), GovReport~\citep{huang-etal-2021-efficient} (long-form government summarization), MultiFieldQA~\citep{bai-etal-2024-longbench} (professional factoid extraction), and LongBench-Chat~\citep{bai-etal-2024-longalign} (real-world conversational QA). This diverse, long-context nature makes it ideal for studying citation granularity effects. Across all model/granularity combinations, our experimental pipeline generated a total of \textbf{60,976 statements}, providing a robust sample size for stratified analysis (\S\ref{subsec:volume-controlled}).

\begin{figure}[t]
    \centering
    \includegraphics[width=\textwidth]{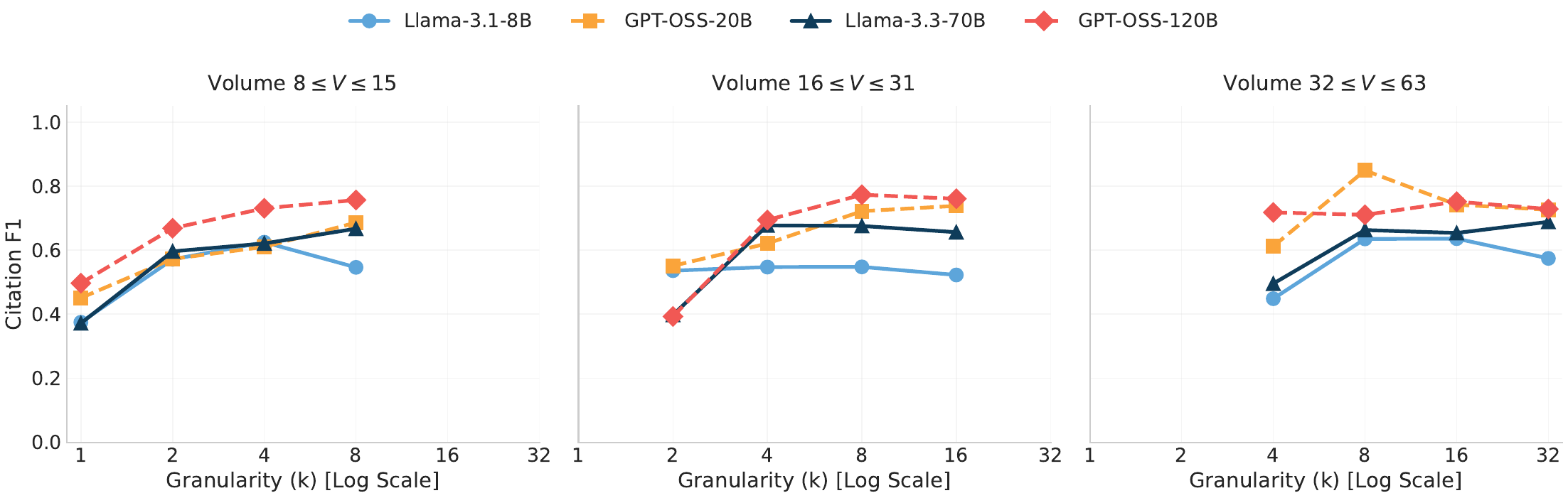}
    \caption{\textbf{The Granularity-Performance Curves.} We track Citation F1 across granularity settings for three representative citation volume ranges. \textit{Across all four evaluated model scales, performance is consistently lowest at fine granularity ($k{=}1,2$) and peaks at intermediate settings before declining or plateauing at coarse settings}.}
    \label{fig:main_phenomenon}
\end{figure}

\subsection{Evaluation Metrics}
\label{subsec:metrics}

We report \textbf{macro-averaged} metrics, assigning equal weight to each dataset to prevent larger datasets from skewing the results. The aggregate attribution and correctness summaries in \S\ref{sec:correctness} are instead computed at the statement level over all generated statements, including those the model left uncited.

\paragraph{Attribution Quality.}
Following the evaluation protocol defined in Section 2.3.2 of \citet{zhang2024longcite}, we evaluate attribution quality at the statement level using an automated judge. 
We replace the proprietary evaluator used in the original work with Qwen3-Next-80B-A3B-Instruct~\citep{yang2025qwen3technicalreport}, while maintaining the same scoring rubric and evaluation prompts. To ensure the reliability of this proxy, we rigorously validated our judge against direct human annotations (achieving 81.0\% exact agreement), GPT-4o outputs ($r=0.92$), and a secondary Llama-3.3-70B evaluator ($r=0.88$). This multi-scale validation is presented in \S\ref{sec:judge_validation}.

For each generated statement $S$ and associated set of cited chunks $\{c_i\}_{i \in I_S}$, the judge computes:
\begin{itemize}[leftmargin=*,noitemsep,topsep=2pt]
    \item \textbf{Citation Precision:} The fraction of cited chunks 
containing information relevant to the statement. Formally, this is 
$|\{c_i\}_{i \in I_S} : c_i \text{ supports } S\}| / |I_S|$ (prompt 
in Table~\ref{tab:prompt_precision}, Appendix~\ref{app:prompt}). This quantifies the \textbf{informational density} of the citations.

    \item \textbf{Citation Recall:} A ternary judgment ($\{0, 0.5, 1\}$) determining whether the \textit{union} of all cited chunks fully supports the statement ($1$), partially supports it ($0.5$), or fails to support it ($0$). This evaluates the \textbf{evidential sufficiency} of the citation set (prompt in Table~\ref{tab:prompt_recall}, Appendix~\ref{app:prompt}).  {Scoring the union makes recall \textbf{chunking-invariant by construction}, complementing precision and isolating genuine evidential coverage from chunk-counting artifacts (\S\ref{sec:mechanism}).}

    \item \textbf{Citation F1:} The harmonic mean of Citation Precision and Citation Recall, computed per statement. Based on these metrics, we define \textbf{optimal granularity} as the specific setting $k$ that maximizes the Citation F1 score for a given model and volume range. Framed as an \textbf{oracle} upper bound, this allows us to identify the specific citation granularity at which a model most effectively balances precision and recall to ground its outputs. {We provide a mechanistic account of why such an optimum must exist in \S\ref{sec:mechanism}.}
\end{itemize}

\paragraph{Answer Correctness.}
Following \citet{zhang2024longcite}, we evaluate factual correctness at the \textbf{query level}. For each query, we concatenate all generated statements into a complete answer, and the Qwen3-Next-80B judge assigns a correctness score based on whether the complete answer is semantically equivalent to the ground-truth reference answer (prompts in Table~\ref{tab:prompt_correctness}, Appendix~\ref{app:prompt}).


\section{Results}\label{sec:results}

\subsection{Performance Peaks at Intermediate Granularity}\label{sec:main_finding}

Across all models and citation volumes, citation F1 exhibits a consistent pattern (Figure~\ref{fig:main_phenomenon}): scores are lowest at very fine granularity ($k{=}1$--2), peak at intermediate granularity ($k{=}4$--8), and stabilize or decline at very coarse granularity ($k{=}16$--32). The optimum is strictly coarser than the finest available setting in all 16 model--volume configurations (sign test $p = 3.1\times10^{-5}$). It lies at $k{=}4$--8 in 12 of the 16 and at $k{=}16$--32 in the remaining four, which cluster at higher volumes. Table~\ref{tab:volume_defense} quantifies the performance gap for Llama-70B, with intermediate granularities consistently outperforming fine-grained baselines by +16--79\%. Critically, fine-grained settings often cite \textit{more} sentences yet achieve lower F1 scores (e.g., 4.9 vs 4.0 sentences at $4 \le V \le 7$), demonstrating that the utility of evidence depends more on its semantic coherence than its raw volume. This parallels findings in RAG systems where increasing fragmentation (while holding total length constant) degrades performance~\citep{levy2025documentslengthisolatingchallenge}, suggesting that fine-grained citation imposes a processing penalty independent of information quantity. (Complete set of performance curves across all volume ranges for all models is provided in Appendix Figure~\ref{fig:app_phenomenon_others}).

\begin{table}[!ht]
\centering
\small
\setlength{\tabcolsep}{4pt}
\begin{tabular}{@{}lccccccccc@{}}
\toprule
\multirow{2}{*}{\textbf{Volume ($V$)}} & \multicolumn{4}{c}{\textbf{Fine-Grained Baseline}} & \multicolumn{4}{c}{\textbf{Optimal Granularity}} & \multirow{2}{*}{\textbf{Gain}} \\
\cmidrule(lr){2-5} \cmidrule(lr){6-9}
& $k$ & F1 & 95\% CI & \textit{N} / Vol & $k$ & F1 & 95\% CI & \textit{N} / Vol & \\
\midrule
$2 \le V \le 3$   & 1 & 0.506 & [0.376, 0.655] & 1253 / 2.3 & 2  & \textbf{0.588} & [0.437, 0.734] & 852 / 2.0  & \textbf{+16.2\%} \\
$4 \le V \le 7$   & 1 & 0.507 & [0.309, 0.657] & 379 / 4.9  & 4  & \textbf{0.687} & [0.526, 0.851] & 1037 / 4.0 & \textbf{+35.5\%} \\
$8 \le V \le 15$  & 1 & 0.372 & [0.176, 0.547] & 339 / 9.8  & 8  & \textbf{0.667} & [0.466, 0.881] & 1077 / 8.0 & \textbf{+79.3\%} \\
$16 \le V \le 31$ & 2 & 0.398 & [0.220, 0.614] & 225 / 19.1 & 4  & \textbf{0.677} & [0.543, 0.819] & 300 / 19.0 & \textbf{+70.1\%} \\
$32 \le V \le 63$\textsuperscript{†} & 4 & 0.495 & [0.398, 0.645] & 173 / 38.2 & 32 & \textbf{0.688} & [0.494, 0.877] & 1495 / 32.0 & \textbf{+39.0\%} \\
\bottomrule
\end{tabular}
\caption{\textbf{Volume-Controlled Attribution Analysis (Llama-70B).} Comparison of Citation F1 at fine-grained versus optimal granularities within fixed volume ranges. 95\% Confidence Intervals via bootstrap over the four datasets. \textbf{Vol}: mean sentences cited; note that the baseline cites at least as much text as the optimum in every range. \textsuperscript{†}$k{=}4$ serves as the baseline because fine-grained citations rarely reach this volume. \textit{Across all volume regimes, enforcing fine-grained citations depresses attribution quality, while intermediate granularities unlock substantial F1 gains.}}
\label{tab:volume_defense}
\end{table}

\subsection{Asymmetric Sensitivity: Attribution Quality vs. Answer Correctness}
\label{sec:correctness}

\begin{figure}[t]
    \centering
    \begin{subfigure}{0.48\linewidth}
        \includegraphics[width=\linewidth]{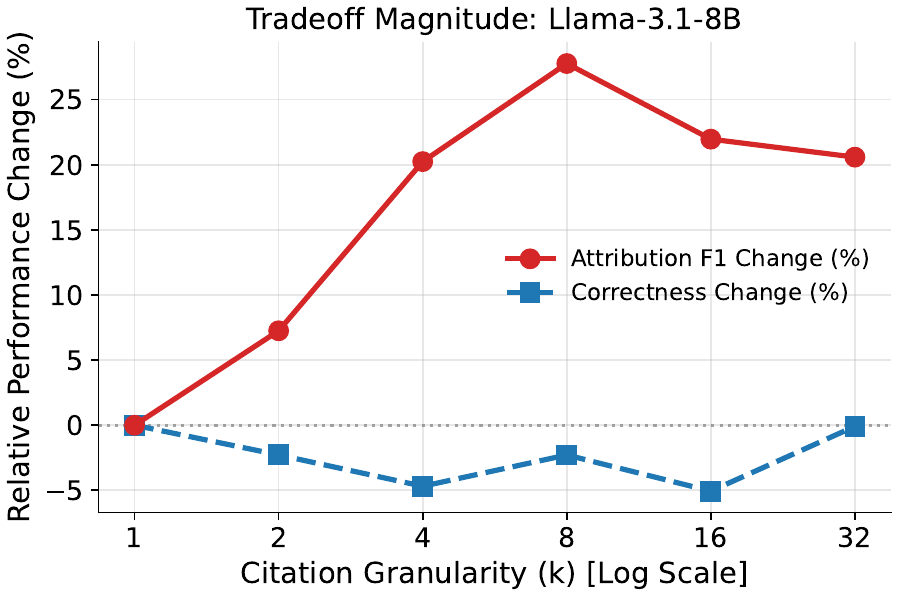}
        \caption{Llama-8B}
    \end{subfigure}
    \hfill
    \begin{subfigure}{0.48\linewidth}
        \includegraphics[width=\linewidth]{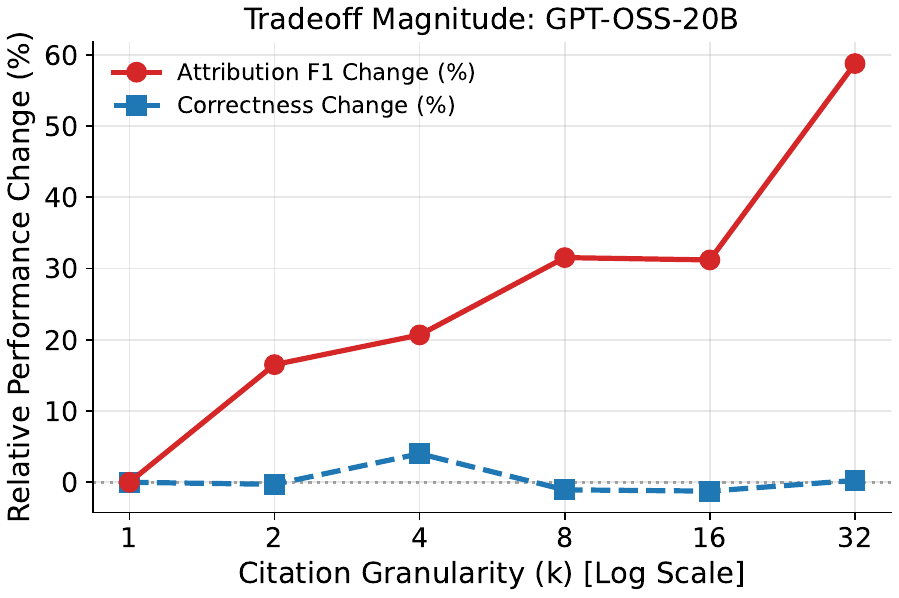}
        \caption{GPT-20B}
    \end{subfigure}

    \vspace{0.8em}

    \begin{subfigure}{0.48\linewidth}
        \includegraphics[width=\linewidth]{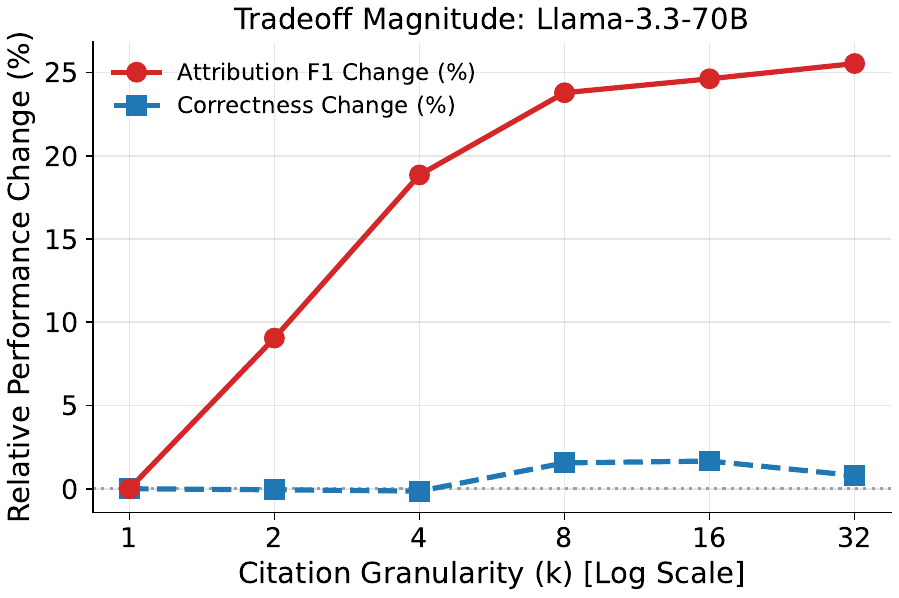}
        \caption{Llama-70B}
    \end{subfigure}
    \hfill
    \begin{subfigure}{0.48\linewidth}
        \includegraphics[width=\linewidth]{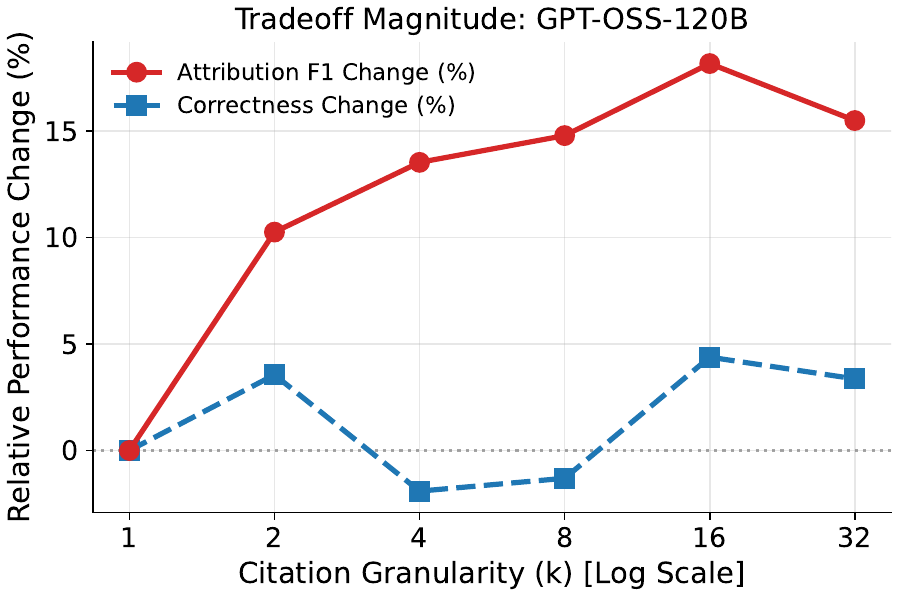}
        \caption{GPT-120B}
    \end{subfigure}
    \caption{\textbf{Asymmetric Sensitivity (all four models).} Lines depict relative change from the fine-grained baseline ($k{=}1$). \textit{Attribution quality (Citation F1, red) responds strongly to granularity, whereas answer correctness (blue) is comparatively flat.} Llama-70B is the clearest case ($>25\%$ attribution gain against $<1\%$ correctness change), and the asymmetry holds across all four scales.}
    \label{fig:correctness_asymmetry}
\end{figure}

Strikingly, the substantial attribution penalty documented above comes at essentially no cost to answer correctness (Figure~\ref{fig:correctness_asymmetry}). We term this decoupling \textbf{Asymmetric Sensitivity}: attribution quality varies dramatically with citation granularity, yet the model's factual reasoning remains stable regardless of how the evidence is 
segmented. This pattern holds across all four models, with 
Table~\ref{tab:asymmetry}a quantifying the effect. For
capable models, the leverage is substantial—Llama-70B achieves a 
25.5\% gain in attribution for a negligible +0.8\% shift in accuracy. Even where trade-offs occur, the exchange remains disproportionate: Llama-8B yields a 27.8\% attribution gain for only a 2.3\% correctness penalty.

\begin{table}[!ht]
\centering
\small
\setlength{\tabcolsep}{6pt}
\begin{tabular}{@{}lccccrr@{}}
\toprule
& \multicolumn{2}{c}{\textbf{Fine-Grained Baseline}} & \multicolumn{2}{c}{\textbf{Citation-Optimal}} & \multirow{2}{*}{\textbf{Cit. Gain}} & \multirow{2}{*}{\textbf{$\Delta$ Acc}} \\
\cmidrule(lr){2-3} \cmidrule(lr){4-5}
& $k$ & Acc & $k$ & Acc & & \\
\midrule
\multicolumn{7}{@{}l}{\textit{(a) Aggregated over all volume ranges, per model}} \\
Llama-8B  & 1 & 0.631 & 8  & 0.617 & +27.8\% & $-2.3$\% \\
GPT-20B   & 1 & 0.564 & 32 & 0.566 & +58.8\% & $+0.3$\% \\
Llama-70B & 1 & 0.719 & 32 & \textbf{0.724} & +25.5\% & $+0.8$\% \\
GPT-120B  & 1 & 0.590 & 16 & \textbf{0.616} & +18.2\% & $+4.4$\% \\
\midrule
\multicolumn{7}{@{}l}{\textit{(b) Llama-70B, within fixed volume ranges}} \\
$4 \le V \le 7$   & 1 & 0.689 & 4  & \textbf{0.730} & +35.5\% & $+5.9$\% \\
$8 \le V \le 15$  & 1 & 0.677 & 8  & \textbf{0.739} & +79.3\% & $+9.2$\% \\
$16 \le V \le 31$ & 2 & 0.676 & 4  & \textbf{0.692} & +70.1\% & $+2.3$\% \\
$32 \le V \le 63$ & 4 & 0.711 & 32 & \textbf{0.716} & +39.0\% & $+0.7$\% \\
\bottomrule
\end{tabular}
\caption{\textbf{Asymmetric Sensitivity.} Citation gain and the corresponding shift in answer correctness, moving from the finest granularity available to the citation-optimal one. \textit{The citation-optimal granularity yields double-digit attribution gains (+18\% to +59\%) with minimal impact on correctness in aggregate (a), and for Llama-70B correctness improves in every volume range (b).} Bold marks an accuracy improvement over the baseline. The correctness-maximizing $k$ (1, 4, 16, and 16 for the four models) seldom coincides with the citation-optimal one, which is what makes granularity separately tunable. Note that (a) pools all volume ranges, so its optimum sits coarser than the within-volume optima of \S\ref{sec:main_finding}, which concentrate at $k{=}4$--8.}
\label{tab:asymmetry}
\end{table}

\paragraph{Correctness stability holds across all citation volumes.}
Table~\ref{tab:asymmetry}b demonstrates that for Llama-70B, this stability holds even as citation volume increases. In every volume regime for this model, the citation-optimal setting yields higher accuracy than the fine-grained baseline. Per-volume correctness for all four models is reported in Appendix~\ref{app:correctness}.


\paragraph{Implications.}
These findings refute fears that relaxing granularity encourages imprecise attribution. Rather than preventing hallucination, fine-grained constraints impose severe quality penalties with negligible correctness benefits. For capable models, relaxing granularity enables more effective verification without compromising factual accuracy.

\subsection{Mechanistic Explanation}\label{sec:mechanism}

\begin{figure}[t]
    \centering
    \begin{subfigure}{0.48\linewidth}
        \includegraphics[width=\linewidth]{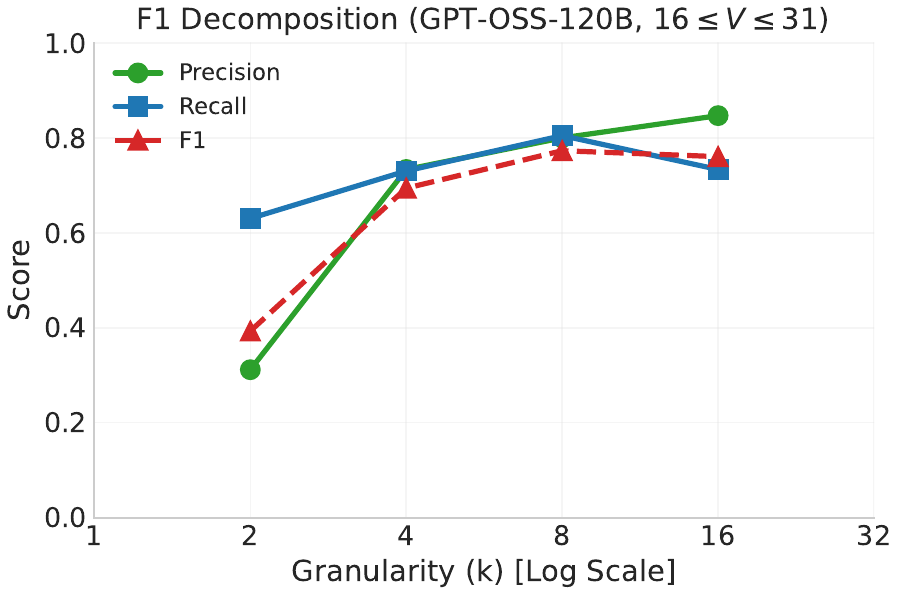}
        \caption{GPT-120B}
    \end{subfigure}
    \hfill
    \begin{subfigure}{0.48\linewidth}
        \includegraphics[width=\linewidth]{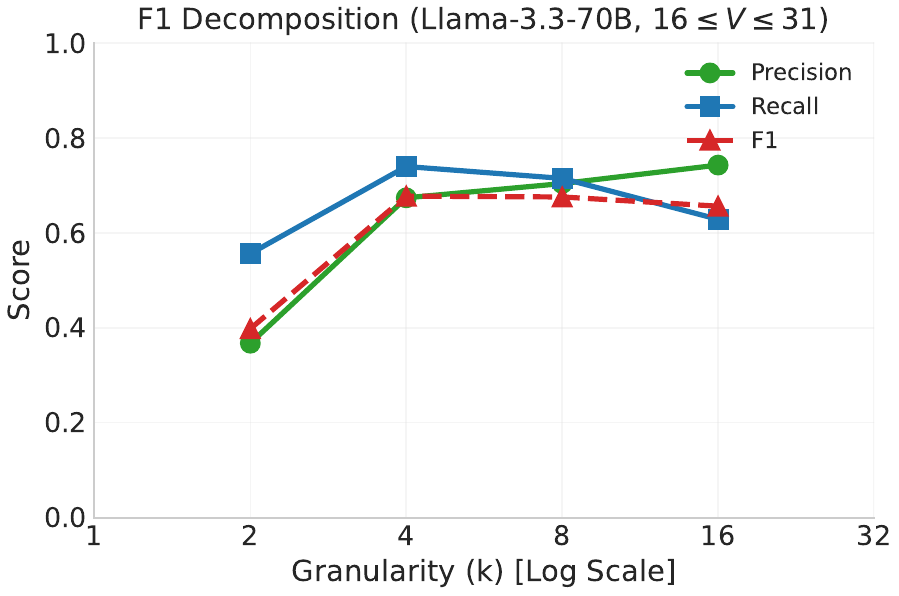}
        \caption{Llama-70B}
    \end{subfigure}
    \caption{\textbf{Citation F1 Decomposition ($\boldsymbol{16 \le V \le 31}$).} \textit{F1 (red) peaks by balancing opposing trends: Precision (green) improves with granularity (better context), while Recall (blue) degrades at coarse settings due to noise.} Shown for the two largest models; the same decomposition for all models and volume ranges is in Appendix~\ref{app:precision_recall}.}
    \label{fig:mechanism}
\end{figure}

We decompose F1 into precision and recall (Figure~\ref{fig:mechanism}, GPT-120B and Llama-70B at $16 \le V \le 31$). {The recall component, being chunking-invariant by construction (\S\ref{subsec:metrics}), provides bias-neutral evidence that coarsening is not uniformly beneficial: recall peaks below the coarsest available setting in 94\% (15/16) of examined model-volume configurations, and declines at the coarsest step in 14/16 (both exceptions lie in $4 \le V \le 7$, where only $k{=}1$--4 are available) (Appendix~\ref{app:precision_recall}).}

\paragraph{Precision: Boundary Tolerance.}
Precision increases monotonically with $k$ (Figure~\ref{fig:mechanism}). For GPT-120B at $16 \le V \le 31$, precision rises from 0.31 ($k{=}2$) to 0.85 ($k{=}16$). We identify this as \textbf{Boundary Tolerance}. While models often successfully locate the semantic \textit{region} of evidence, they frequently struggle to pinpoint exact syntactic start/end points~\citep{wu2024retrievalheadmechanisticallyexplains}. Coarser granularity mitigates this issue by absorbing localization noise, effectively converting ``near-miss'' citations into valid hits.

\paragraph{Recall: Signal Dilution at Coarse Scales.}
Counter-intuitively, broader context does not guarantee better coverage. Recall peaks at intermediate granularities ($k{=}8$) before degrading at coarse settings ($k{=}16$). This provides empirical evidence of \textbf{Signal Dilution}: as chunks expand beyond the boundaries of the relevant evidence, the ratio of irrelevant to relevant tokens increases, overwhelming the model's attention mechanism and causing it to miss evidence it previously found~\citep{pmlr-v202-shi23a,liu2023lost}.

\paragraph{The Optimal Balance.}
The peak performance at intermediate granularities emerges from this trade-off: at fine granularities, low precision dominates; at coarse granularities, declining recall dominates; at intermediate granularities ($k{=}4$--8 in 75\% of cases), the trade-off balances optimally. This identifies the citation granularity that best matches the model's semantic processing scale—representing the natural scope within which it synthesizes information most effectively.

\paragraph{Is the Pattern a Metric Artifact?}
Our Citation Precision metric follows the standard LongBench-Cite protocol~\citep{zhang2024longcite}, counting a chunk as correct if it contains any relevant information. This definition is structurally more lenient at coarser granularities, where irrelevant sentences may be absorbed within an otherwise valid block~\citep{qian2026relevant}, so it is worth asking whether the inverted-U is produced by the metric rather than by the models. Two pieces of evidence say it is not. First, recall is chunking-invariant by construction (\S\ref{subsec:metrics}) and is therefore immune to this leniency, yet it does not rise monotonically with $k$: it peaks below the coarsest available setting in 15 of 16 configurations and falls at the coarsest step in 14 of 16 (Appendix~\ref{app:precision_recall}). Leniency can inflate precision, but it cannot make recall decline. Second, the precision pattern is itself diagnostic. At sentence-level granularity, citing the single most relevant sentence would maximize precision, yet models consistently include additional sentences and accept the penalty. This indicates that models reliably locate the relevant semantic region but operate at a scope broader than a single sentence---making sentence-level precision an unnatural target rather than a metric artifact.

\subsection{Scale-Dependent Granularity Effects}\label{sec:scale}

While the preference for coarser-than-finest granularity holds in all 16 model--volume configurations, the \textit{magnitude} of performance sensitivity varies substantially across model scales, and does so in a volume-dependent way. 
Table~\ref{tab:scale_sensitivity} summarizes the relative F1 gains moving from fine-grained baselines to optimal settings.



\begin{table}[!ht]
\centering
\small
\setlength{\tabcolsep}{4pt}
\begin{tabular}{@{}lccccccccc@{}}
\toprule
\multirow{2}{*}{\textbf{Volume ($V$)}} & \multirow{2}{*}{\textbf{Base $k$}} & \multicolumn{2}{c}{\textbf{Llama-8B}} & \multicolumn{2}{c}{\textbf{GPT-20B}} & \multicolumn{2}{c}{\textbf{Llama-70B}} & \multicolumn{2}{c}{\textbf{GPT-120B}} \\
\cmidrule(lr){3-4} \cmidrule(lr){5-6} \cmidrule(lr){7-8} \cmidrule(l){9-10}
& & Opt $k$ & Gain & Opt $k$ & Gain & Opt $k$ & Gain & Opt $k$ & Gain \\
\midrule
\textbf{4--7}   & 1 & 4  & +22.4\% & 4  & +28.8\% & 4  & +35.4\% & 4  & \textbf{+47.9\%} \\
\textbf{8--15}  & 1 & 4  & +66.7\% & 8  & +52.2\% & 8  & \textbf{+79.5\%} & 8  & +52.3\% \\
\textbf{16--31} & 2 & 8  & +2.2\%  & 16 & +34.0\% & 4  & +70.2\% & 8  & \textbf{+96.7\%} \\
\textbf{32--63}\textsuperscript{†} & 4 & 16 & \textbf{+41.8\%} & 8  & +38.7\% & 32 & +38.9\% & 16 & +4.7\%  \\
\bottomrule
\end{tabular}
\caption{\textbf{Scale-Dependent Granularity Effects.} Relative F1 gains from the finest granularity available in each volume range to the optimal setting; the largest gain in each row is bold. \textit{Gains rise with model size in the $4$--$7$ and $16$--$31$ ranges, indicating that fine-grained constraints disproportionately restrict the synthesis capabilities of larger models.} \textsuperscript{†}$k{=}4$ baseline used due to sample scarcity. Absolute baseline F1 values are given in Appendix~\ref{app:scale_effects}, Table~\ref{tab:scale_sensitivity_complete}.}
\label{tab:scale_sensitivity}
\end{table}

\paragraph{Progressive Sensitivity and the Semantic Gap ($4 \le V \le 31$).}
In low-volume regimes ($4 \le V \le 7$), benefits scale with model size (+22.4\% for 8B vs. +47.9\% for 120B). While the $8 \le V \le 15$ range shows fluctuation, the critical divergence appears in the $16 \le V \le 31$ range. Here, the 8B model is nearly insensitive (+2.2\% gain), whereas larger models show massive sensitivity: 70B gains +70.2\% and 120B gains +96.7\% (improving from 0.393 at $k{=}2$ to 0.773 at $k{=}8$). We hypothesize this occurs because larger models possess stronger semantic synthesis capabilities~\citep{chen2025surveyscalinglargelanguage,wei2022emergent} that they cannot effectively leverage under fine-grained constraints.

\paragraph{High-Volume Saturation ($32 \le V \le 63$).}
At high volumes, the 120B model shows minimal gain (+4.7\%), likely because the baseline is already moderate ($k{=}4$). This suggests when a capable model voluntarily aggregates this much evidence, it effectively integrates information across the many distinct segments (e.g., 8 chunks at $k{=}4$) without performance loss. Other models still gain +38--42\% from coarser settings, indicating they struggle to coordinate this high chunk count; for them, consolidating evidence into fewer blocks restores the structural simplicity required for synthesis.

\paragraph{Implications.}
Current benchmarks enforcing fine-grained citation systematically underestimate advanced model capabilities. Because citation sensitivity scales with model size in specific volume regimes, metrics that enforce sentence-level granularity disproportionately penalize the semantic processing strengths of large models (70B+).

\subsection{Task-Dependent Optimal Granularity}\label{sec:generalization}

The specific attribution-optimal granularity depends on the nature of the task, as shown by the performance of Llama-70B in the representative $16 \le V \le 31$ volume range detailed in Table~\ref{tab:per_dataset_full}.

\begin{table}[!ht]
\centering
\small
\setlength{\tabcolsep}{4pt}
\begin{tabular}{@{}lcccccc@{}}
\toprule
\textbf{Dataset} & \textbf{Type} & \multicolumn{4}{c}{\textbf{Citation F1}} & \textbf{Opt} \\
\cmidrule(lr){3-6}
& & $k{=}2$ & $k{=}4$ & $k{=}8$ & $k{=}16$ & \\
\midrule
HotpotQA & Multi-hop & 0.315 & \textbf{0.602} & 0.565 & 0.459 & 4 \\
LB-Chat & Chat/QA & 0.129 & \textbf{0.485} & 0.482 & 0.443 & 4 \\
GovReport & Summ. & 0.653 & 0.826 & 0.879 & \textbf{0.901} & 16 \\
MultiField & Factoid & 0.494 & 0.795 & 0.775 & \textbf{0.821} & 16 \\
\midrule
\textbf{Macro} & --- & 0.398 & \textbf{0.677} & 0.675 & 0.656 & 4 \\
\bottomrule
\end{tabular}
\caption{\textbf{Task-Dependent Optimal Granularity (Llama-70B, $\boldsymbol{16 \le V \le 31}$).} \textit{Reasoning tasks (HotpotQA) peak at intermediate $k{=}4$, whereas summarization (GovReport) favors coarser context ($k{=}16$). }All domains outperform the fine-grained baseline ($k{=}2$).}
\label{tab:per_dataset_full}
\end{table}

\paragraph{Task requirements drive optima.}
We observe a shift in optimal granularity driven by task requirements. Summarization tasks like GovReport benefit from coarser granularity ($k{=}16$), allowing the model to synthesize information from larger contiguous blocks. In contrast, multi-hop QA tasks like HotpotQA require isolating specific details while maintaining the links between them, resulting in an intermediate granularity ($k{=}4$) that balances sufficient context to bridge reasoning steps with minimal noise from distractors. The driver appears to be the spatial distribution of the evidence rather than reasoning-chain length alone: in GovReport the supporting evidence forms long contiguous blocks that coarse chunks preserve intact, whereas in HotpotQA it consists of distant discrete facts plus the bridge between them, so $k{=}4$ keeps each fact with its local context while excluding the intervening distractors.

\paragraph{Fine granularity is consistently suboptimal.}
Despite these differing optima, most citation volumes across all datasets show large gains over the fine granularity. Even for QA tasks requiring specific information, $k{=}4$ outperforms $k{=}2$ by substantial margins (e.g., +91\% on HotpotQA). Across all models, gains over $k{=}2$ range from 0\% to 338\% (Appendix~\ref{app:tasks}, Table~\ref{tab:dataset_comprehensive}), confirming that atomic sentence-level citation is largely suboptimal.

\subsection{From Diagnosis to Policy: Task-Calibrated Granularity}\label{sec:policy}

The analysis so far identifies the optimal granularity in hindsight, which raises the question of whether the finding is actionable or merely diagnostic. Because the optimum is largely a property of the task (\S\ref{sec:generalization}), a single parameter per task is enough to recover most of it. We split the 585 queries 50/50, stratified by dataset, select the Citation F1-maximizing $k$ per dataset on one half, and apply it unchanged to the held-out half; the baseline fixes $k{=}1$ and the oracle selects the best $k$ per dataset directly on the held-out half. Results average five random splits with no query overlap, macro-averaged over the four datasets as elsewhere.

Table~\ref{tab:policy} reports the outcome. A one-parameter-per-task policy raises Citation F1 by 16--30\% (mean $+23.2\%$) and captures 64--87\% (mean 73\%) of the oracle gain, with the same ordering under the analyzed volume range $2 \le V \le 63$ (74\% capture, $+29\%$). The policy selects $k$ per \textit{task}, so it presumes the task type is known at inference time; selecting $k$ per query, which would relax that assumption, is a natural next step. Even in this simplest form, the result establishes that the granularity effect we document is not only measurable but recoverable at negligible cost.

\begin{table}[!ht]
\centering
\small

\begin{tabular}{@{}lcccrr@{}}
\toprule
\textbf{Model} & \textbf{Base F1 ($k{=}1$)} & \textbf{Policy} & \textbf{Oracle} & \textbf{Gain} & \textbf{Captured} \\
\midrule
Llama-8B    & 0.472 & 0.579 & 0.617 & +23.1\% & 72.3\% \\
GPT-20B     & 0.596 & 0.734 & 0.797 & +23.5\% & 69.2\% \\
Llama-70B   & 0.520 & 0.675 & 0.697 & +30.0\% & 87.4\% \\
GPT-120B    & 0.631 & 0.732 & 0.787 & +16.1\% & 64.0\% \\
\midrule
\textbf{Mean} & --- & --- & --- & \textbf{+23.2\%} & \textbf{73.3\%} \\
\bottomrule
\end{tabular}
\caption{\textbf{Task-Calibrated Granularity Selection.} Selecting one $k$ per dataset on a held-out calibration half recovers most of the oracle gain over the sentence-level baseline. Averaged over five random 50/50 splits with no query overlap; F1 macro-averaged over the four datasets.}
\label{tab:policy}
\end{table}

\section{Validation of Automated Evaluation}
\label{sec:judge_validation}

We validate the Qwen3-80B judge at three escalating scales: direct human evaluation, correlation with a proprietary model previously validated against human experts, and large-scale consistency against an alternative open-weights judge. All three are stratified by granularity, so that any granularity-dependent evaluator bias would be visible.

\paragraph{Direct Human Evaluation (200 Samples).}
We conducted a blind human evaluation of 200 sampled statements, annotated by the authors. The sample was balanced across datasets, models, and granularity levels ($k{=}1$ to $k{=}32$), and annotators were blinded to all three. Evaluating one statement at a time, annotators completed two steps:
\begin{itemize}[leftmargin=*,noitemsep,topsep=2pt]
    \item \textbf{Precision:} Annotators labeled each cited chunk individually as \textit{Relevant} or \textit{Irrelevant}. The statement's overall precision is the average of these chunk-level scores.
    \item \textbf{Recall:} Following the strict evaluation protocols of LongBench-Cite~\citep{zhang2024longcite} and ALCE~\citep{gao2023enabling}, which treat ``partial support'' as ``no support,'' annotators evaluated the statement against all cited chunks combined to give a strict binary score: \textit{Fully Supported} or \textit{No Support}.
\end{itemize}

As shown in Table~\ref{tab:human_eval}, Qwen3-80B agrees closely with human judgment: Citation F1 reaches 81.0\% exact agreement with Cohen's $\kappa = 0.706$. Critically for our purposes, this alignment is stable across granularity bands---82.9\%, 79.7\%, and 80.3\% exact agreement at fine, intermediate, and coarse settings---so the judge does not systematically favor one chunk size over another.

\paragraph{GPT-4o Correlation (600 Samples).}
We next compare Qwen3-80B against GPT-4o, the evaluator that \citet{zhang2024longcite} validated against human experts (Cohen's $\kappa \approx 0.60$). On 600 generated statements stratified evenly across all four models and all six granularities, we observe high alignment on every metric: Citation F1 (Pearson $r = 0.92$, $\kappa = 0.79$), Precision ($r = 0.88$, $\kappa = 0.80$), and Recall ($r = 0.84$, $\kappa = 0.71$). Qwen3-80B thus reproduces GPT-4o's judgments closely enough to inherit its established human alignment at a scale we could not annotate by hand.

\paragraph{Large-Scale Consistency (35,926 Statements).}
Finally, to test robustness to the choice of evaluator, we re-scored with a second open-weights judge, Llama-3.3-70B. We re-evaluated all 35,926 statements generated by Llama-3.1-8B and GPT-OSS-120B---the smallest and largest models in our study---across all six granularity settings. Agreement is high and, more importantly, flat: Citation F1 correlates at $r = 0.877$ overall with negligible bias ($+0.006$), and per-granularity correlation stays within $r = 0.868$--$0.887$ from $k{=}1$ to $k{=}32$, with no degradation at either extreme (Table~\ref{tab:judge_validation_wide}). The granularity-performance curve we report is therefore a property of the citations, not of a particular judge.

\begin{table}[ht]
\centering
\small
\setlength{\tabcolsep}{6pt} 
\begin{tabular}{llccc}
\toprule
\textbf{Metric} & \textbf{Granularity ($k$)} & \textbf{Pearson $r$ (\%)} & \textbf{Cohen’s $\kappa$ (\%)} & \textbf{Agreement (\%)} \\
\midrule
\multirow{4}{*}{Precision} 
& \textbf{Overall (All $k$)} & \textbf{73.2} & \textbf{66.0} & \textbf{80.5} \\
& \hspace{0.5em} Fine (1, 2)    & 74.9 & 66.7 & 77.1 \\
& \hspace{0.5em} Interm. (4, 8) & 74.6 & 70.7 & 84.4 \\
& \hspace{0.5em} Coarse (16, 32)& 69.1 & 59.6 & 80.3 \\
\midrule
\multirow{4}{*}{Recall} 
& \textbf{Overall (All $k$)} & \textbf{65.2} & \textbf{64.0} & \textbf{82.5} \\
& \hspace{0.5em} Fine (1, 2)    & 71.6 & 71.1 & 85.7 \\
& \hspace{0.5em} Interm. (4, 8) & 65.8 & 64.1 & 82.8 \\
& \hspace{0.5em} Coarse (16, 32)& 57.5 & 55.6 & 78.8 \\
\midrule
\multirow{4}{*}{\textbf{Citation F1}} 
& \textbf{Overall (All $k$)} & \textbf{72.3} & \textbf{70.6} & \textbf{81.0} \\
& \hspace{0.5em} Fine (1, 2)    & 81.6 & 77.9 & 82.9 \\
& \hspace{0.5em} Interm. (4, 8) & 69.7 & 68.5 & 79.7 \\
& \hspace{0.5em} Coarse (16, 32)& 64.9 & 63.8 & 80.3 \\
\bottomrule
\end{tabular}
\caption{\textbf{Human-to-LLM (Qwen3-80B) Agreement Metrics.} Across all levels of granularity ($k$), the automated judge exhibits substantial agreement with human annotators (Citation F1: $\kappa = 0.64$--$0.78$) and exact agreement of at least 77\% on every metric and granularity band. This stability confirms that the evaluator is a robust and unbiased proxy for both fine-grained and coarse-grained attribution quality.}
\label{tab:human_eval}
\end{table}

\begin{table}[t]
\centering
\small
\setlength{\tabcolsep}{7pt}
\begin{tabular}{@{}lrccc@{}}
\toprule
\textbf{Granularity} & \textbf{Samples ($N$)} & \textbf{Precision} & \textbf{Recall} & \textbf{Citation F1} \\
\midrule
$k=1$  & 5,907 & 0.824 & 0.896 & 0.884 \\
$k=2$  & 6,000 & 0.831 & 0.885 & 0.880 \\
$k=4$  & 5,958 & 0.864 & 0.878 & 0.887 \\
$k=8$  & 6,047 & 0.846 & 0.856 & 0.875 \\
$k=16$ & 6,056 & 0.844 & 0.852 & 0.868 \\
$k=32$ & 5,958 & 0.846 & 0.863 & 0.877 \\
\midrule
\textbf{All granularities} & \textbf{35,926} & \textbf{0.841} & \textbf{0.871} & \textbf{0.877} \\
\textit{Bias (mean difference)} & & $-0.010$ & $+0.023$ & $+0.006$ \\
\bottomrule
\end{tabular}
\caption{\textbf{Judge Consistency Analysis.} Pearson correlation between our primary judge (Qwen3-Next-80B) and a secondary judge (Llama-3.3-70B). Agreement on Citation F1 is high and, critically, flat across chunk sizes ($r = 0.868$--$0.887$ from $k{=}1$ to $k{=}32$), with no degradation at either extreme---the performance curve we report is a property of the citations rather than of a particular evaluator. Precision and Recall are stable over the same range ($r = 0.824$--$0.864$ and $r = 0.852$--$0.896$). Systematic bias is negligible on every metric and all correlations are significant at $p < 0.001$.}
\label{tab:judge_validation_wide}
\end{table}

\section{Discussion}

\paragraph{Granularity Mismatch and Asymmetric Sensitivity.}
We attribute the failure of fine-grained citation to a \textbf{Granularity Mismatch}: a fundamental conflict between imposed sentence-level constraints and the model's effective semantic scope~\citep{weng-etal-2023-large, arafat2025citationgroundedcodecomprehensionpreventing}, 
with the penalty falling hardest on the most capable models at low and medium citation volumes. Crucially, this cost is paid in attribution alone—an \textbf{Asymmetric Sensitivity} where attribution quality varies substantially while answer correctness remains stable. {The implication is concrete: because granularity
affects attribution but not correctness, it can be tuned independently to improve grounding, and an evaluation reporting only one of the two axes would miss the effect entirely~\citep{sun2026accuracymeasuringbiasacknowledgment}. The field has overlooked this freedom by treating sentence-level citation as a default rather than a deliberate design
choice.}

\paragraph{What Sets the Optimal Granularity.}
Our results on volume, scale, and task are easier to read as one. In each case the best chunk size tracks the span of text that the claim being supported actually depends on. Summarization claims in GovReport rest on long continuous passages, and the optimum there is coarse ($k{=}16$); multi-hop claims in HotpotQA rest on two separate facts plus the sentence linking them, and the optimum is intermediate ($k{=}4$). Claims that need more evidence in total push the optimum coarser still, which is why it moves from $k{=}4$--8 at low citation volumes to $k{=}16$--32 at the highest in most configurations. And larger models, which draw on wider spans of text, lose the most when those spans are cut into single sentences. Read this way, granularity is not a formatting parameter to be set once. It is a standing estimate of how much context a claim needs: set it too fine and the model is denied context it requires, set it too coarse and it is handed text it must then ignore.

\paragraph{Does This Trade Against Human Verifiability?}
Fine-grained citation was adopted because a short citation is easier for a person to check, and we do not measure that benefit directly. Our results still bear on it. The case for sentence-level citation pictures one well-chosen sentence, but that is not what models produce under the constraint. In the $8 \le V \le 15$ range at $k{=}1$, they cite roughly nine separate sentences per statement, of which only 30--39\% are judged relevant. A reader checking such a statement must find nine fragments scattered through the document and discover that most of them do not matter. The appeal to verification is an efficiency argument, but efficiency is second-order: it presupposes that the claim \textit{can} be checked from what was cited at all, and that is settled by whether the model could ground it at that granularity in the first place. Where it could not, optimizing the citation format for speed has produced evidence that is not merely slower to check but harder to check. Our own experience annotating the 200-statement sample points the same way, though we did not measure it: chunk length is apparent to an annotator even when the granularity label is hidden, and at $k{\le}2$ we often had to read beyond the cited sentence to judge a claim, while at $k{\ge}16$ locating the relevant part within the cited block became the work itself. We report this as an impression rather than a result. We therefore regard the trade-off between model attribution and human verification as an open question rather than a settled one, and measuring verification effort directly---success rate and time per item---as the way to answer it.

\paragraph{Granularity Changes Which Model Looks Better.}
Because the penalty is not the same for every model, the chunk size a benchmark fixes can change the conclusion it reports. In the $16 \le V \le 31$ range, ranking the four models by Citation F1 at $k{=}2$ gives GPT-20B, Llama-8B, Llama-70B, GPT-120B. At $k{=}16$ the order is GPT-120B, GPT-20B, Llama-70B, Llama-8B: the model that ranked last now ranks first, and the model that ranked second now ranks last. Five of the six pairwise comparisons in this range change direction somewhere between $k{=}2$ and $k{=}16$, which is visible in Figure~\ref{fig:main_phenomenon} as curves that cross rather than run parallel. A benchmark fixing sentence-level citation would report that Llama-8B attributes more reliably than both Llama-70B and GPT-120B (0.536 against 0.398 and 0.393); at $k{=}4$ both larger models overtake it (0.677 and 0.694 against 0.547). The granularity a benchmark chooses is therefore not a neutral formatting decision, but a configuration its conclusions are conditional on~\citep{li2026safetyreproconfigurationconditionalrankinstability}.

\paragraph{Implications for Attribution Systems.}
A benchmark that fixes one granularity does not simply measure attribution imprecisely; it measures a different quantity than it intends, and can order models wrongly while doing so. We see three directions
forward:
\begin{itemize}[leftmargin=*,noitemsep,topsep=2pt]
    \item \textbf{Multi-Granularity Evaluation:} Benchmarks should 
    evaluate across granularities rather than mandating sentence-level, 
    balancing human verifiability with model semantic scope~\citep{li2026safetyreproconfigurationconditionalrankinstability}.
    \item \textbf{Task-Adaptive Scoring:} Optimal granularity depends 
    on task; benchmarks must support diverse settings rather than 
    imposing a single standard.
    \item \textbf{Adaptive Granularity Selection:} Since the optimum
    varies with model, volume, and task, systems should select
    granularity dynamically—via training signal or inference-time
    control—rather than fixing a uniform standard. Our task-calibrated
    policy (\S\ref{sec:policy}) is a first step, recovering 73\% of the
    oracle gain from a single parameter per task; selecting $k$ per query
    remains open.
\end{itemize}

\section{Conclusion}

{Our analysis exposes a paradox in current attribution design: fine-grained standards, adopted for human verifiability, fracture the semantic dependencies models require to ground claims faithfully. Across four model scales, attribution quality consistently peaks at intermediate (paragraph-level) granularity, and---most strikingly---this gain comes at no measurable cost to answer correctness. At low and medium citation volumes, the penalty falls hardest on the strongest models, whose multi-sentence synthesis the fine-grained constraint disproportionately suppresses. Ultimately, effective attributed generation requires aligning citation units with the model's inherent processing scale to preserve semantic continuity. Citation granularity must therefore be treated as a model-centric parameter to be measured, rather than a fixed standard prescribed by benchmarks.}

\section{Limitations}
\label{sec:limitations}

\paragraph{Reasoning versus Attribution.}
Because answers and citations are produced in a single pass, changing $k$ changes the generated answer as well as its citations. Our metrics consequently entangle the two, and we cannot say whether the degradation at fine granularity reflects disrupted reasoning or disrupted citation alignment---nor whether a model reasoned over the evidence it cites or attached the citation afterward. A post hoc pipeline, in which answers are generated first and citations assigned second, would separate them; we view our joint-generation results as the baseline such work should be measured against.

\paragraph{Retrieval Noise.}
Our results characterize a regime in which the relevant evidence is guaranteed to be present. How granularity effects compose with imperfect retrieval---particularly when retrieved chunks are themselves poorly formed units---is untested here.

\paragraph{Language Scope.}
While LongBench-Cite provides a robust meta-benchmark of diverse tasks and domains, our analysis is restricted to a single language (English). We prioritized this dataset because its extreme document lengths uniquely provide the inputs necessary to evaluate citation behavior across the full range of citation volumes, including high-volume regimes (e.g., $V \ge 32$). Generalization to non-English languages remains an area for future investigation.


\section*{Acknowledgments}
This work is supported in part by Amazon, Defense Advanced Research Projects Agency (DARPA) under Contract No. HR001125C0304 and ONR grant (N0001424-1-2089).
Any opinions, findings and conclusions or recommendations expressed in this material are those of the author(s) and do not necessarily reflect the views of DARPA. 

\bibliography{ref}
\bibliographystyle{tmlr}

\clearpage
\appendix

\section{Prompt Template}
\label{app:prompt}

\subsection{Generation Prompt}
As shown in Table~\ref{tab:prompt_generation}, we used a single unified prompt for all models to minimize format-based bias. \textbf{[CONTEXT]} and \textbf{[QUESTION]} were replaced with the segmented document and query respectively.

\begin{table}[t]
\centering
\small
\begin{tabular}{p{0.95\textwidth}}
\toprule
\textbf{Generation Prompt} \\
\midrule
You are given a question and a document divided into chunks labeled <C0>, <C1>, <C2>, etc.
Your task is to answer the question concisely and support every factual claim with citations.

\vspace{0.5em}
\textbf{Output Format:}\\
- Write your answer using ONLY <statement> elements\\
- Each statement should express a single factual claim\\
- Follow each claim with <cite>[chunk numbers]</cite>\\
- For non-factual statements, use <cite></cite>

\vspace{0.5em}
\textbf{Citation Rules:}\\
- Be precise: cite only chunks that support each claim\\
- Format: <cite>[0-2]</cite> for chunks C0 through C2\\
- Multiple ranges: <cite>[0-2][5-6]</cite>

\vspace{0.5em}
\textbf{Example:}\\
<statement>Earth's atmosphere is primarily nitrogen<cite>[3]</cite></statement>

\vspace{0.5em}
[Document Start]\\
\textbf{[CONTEXT]}\\ {}
[Document End]

\vspace{0.5em}
Question: \textbf{[QUESTION]} \\
\bottomrule
\end{tabular}
\caption{Unified generation prompt template. The granularity of the chunks provided in \textbf{[CONTEXT]} (e.g., 1 sentence vs 16 sentences) is the only variable manipulated.}
\label{tab:prompt_generation}
\end{table}

\subsection{Evaluation Prompts}
Following the established protocol of LongBench-Cite \citep{zhang2024longcite}, we utilize an LLM judge (Qwen3-80B) to evaluate the attribution quality and factual correctness of the generated outputs. 

Table~\ref{tab:prompt_precision} and Table~\ref{tab:prompt_recall} present the exact prompts used to calculate statement-level Citation Precision (relevance) and Citation Recall (support), respectively. Table~\ref{tab:prompt_correctness} presents the query-level Answer Correctness prompts, which are tailored to the specific formatting and scaling requirements of the underlying datasets (e.g., 1--3 scale for QA, 1--5 scale for summarization, and a 1--10 few-shot scale for conversational queries). Variables in bold brackets, such as \textbf{[QUESTION]} or \textbf{[STATEMENT]}, are dynamically replaced with the corresponding inputs for each evaluation step.

\begin{table}[t]
\centering
\small
\begin{tabular}{p{0.95\textwidth}}
\toprule
\textbf{Evaluation Prompt: Citation Precision (Relevance)} \\
\midrule
You are an expert in evaluating text quality. You will receive a user's question about an uploaded document, a factual statement from an AI assistant's response based on that document, and a snippet from the document (since the document is too long to display in full). Your task is to carefully assess whether the snippet contains some key information of the statement. Please use the following grades to generate the rating:

- [[Relevant]] - Some key points of the statement are supported by the snippet or extracted from it.

- [[Unrelevant]] - The statement is almostly unrelated to the snippet, or all key points of the statement are inconsistent with the snippet content.

Ensure that you do not use any information or knowledge outside of the snippet when evaluating. 
Please provide the rating first, followed by the analysis, in the format "Rating: [[...]] Analysis: ...". \vspace{0.5em}

<question>
\textbf{[QUESTION]}
</question>

<statement>
\textbf{[STATEMENT]}
</statement>

<snippet>
\textbf{[CITED CHUNK]}
</snippet> \\
\bottomrule
\end{tabular}
\caption{LLM Judge evaluation prompt used to calculate Citation Precision (Relevance) at the chunk level.}
\label{tab:prompt_precision}
\end{table}

\begin{table}[t]
\centering
\small
\begin{tabular}{p{0.95\textwidth}}
\toprule
\textbf{Evaluation Prompt: Citation Recall (Support)} \\
\midrule
You are an expert in evaluating text quality. You will receive a user's question about an uploaded document, a factual statement from an AI assistant's response based on that document, and a snippet from the document (since the document is too long to display in full). Your task is to carefully assess whether this statement is supported by the snippet. Please use the following scale to generate your rating:

- [[Fully supported]] - Most information in the statement is supported by or extracted from the snippet. This applies only to cases where the statement and parts of the snippet are almost identical. 

- [[Partially supported]] - More than half of the content in the statement is supported by the snippet, but a small portion is either not mentioned or contradicts the snippet. For example, if the statement has two key points and the snippet supports only one of them, it should be considered [Partially supported].

- [[No support]] - The statement is largely unrelated to the snippet, or most key points in the statement do not align with the content of the snippet.

Ensure that you do not use any information or knowledge outside of the snippet when evaluating. 
Please provide the rating first, followed by the analysis, in the format "Rating: [[...]] Analysis: ...". \vspace{0.5em}

<question>
\textbf{[QUESTION]}
</question>

<statement>
\textbf{[STATEMENT]}
</statement>

<snippet>
\textbf{[UNION OF ALL CITED CHUNKS]}
</snippet> \\
\bottomrule
\end{tabular}
\caption{LLM Judge evaluation prompt used to calculate Citation Recall (Support) against the union of all cited chunks.}
\label{tab:prompt_recall}
\end{table}

\begin{table}[t]
\centering
\small
\begin{tabular}{p{0.95\textwidth}}
\toprule
\textbf{Evaluation Prompt: Answer Correctness (MultiFieldQA, HotpotQA)} \\
\midrule
You are asked to evaluate the quality of the AI assistant's answers to user question as an impartial judge, and your evaluation should take into account factors including correctness (high priority), and comprehensiveness (whether the assistant's answer covers all points). Read the AI assistant's answer and compare against the reference answer, and give an overall integer rating in 1, 2, 3 (1 = wrong or irrelevant, 2 = partially correct, 3 = correct and comprehensive) based on the above principles, strictly in the following format: ``[[rating]]'', e.g. ``[[2]]''.

Question:
\textbf{[QUESTION]}

Reference answer:
\textbf{[REFERENCE ANSWER]}

Assistant's answer:
\textbf{[RESPONSE]}

Rating: \\
\midrule
\textbf{Evaluation Prompt: Answer Correctness (GovReport)} \\
\midrule
You are asked to evaluate the quality of the AI assistant's generated summary as an impartial judge, and your evaluation should take into account factors including correctness (high priority), comprehensiveness (whether the assistant's summary covers all points), and coherence. Read the AI assistant's summary and compare against the reference summary, and give an overall integer rating in on a scale of 1 to 5, where 1 is the lowest and 5 is the highest based on the evaluation criteria, strictly in the following format: ``[[rating]]'', e.g. ``[[3]]''.

Question:
\textbf{[QUESTION]}

Reference answer:
\textbf{[REFERENCE ANSWER]}

Assistant's answer:
\textbf{[RESPONSE]}

Rating: \\
\midrule
\textbf{Evaluation Prompt: Answer Correctness (LongBench-Chat)} \\
\midrule
You are asked to evaluate the quality of the AI assistant's answers to user questions as an impartial judge, and your evaluation should take into account factors including correctness (high priority), helpfulness, accuracy, and relevance. The scoring principles are as follows: 1. Read the AI assistant's answer and compare the assistant's answer with the reference answer. 2. Identify all errors in the AI Assistant's answers and consider how much they affect the answer to the question. 3. Evaluate how helpful the AI assistant's answers are in directly answering the user's questions and providing the information the user needs. 4. Examine any additional information in the AI assistant's answer to ensure that it is correct and closely related to the question. If this information is incorrect or not relevant to the question, points should be deducted from the overall score. Please give an overall integer rating from 1 to 10 based on the above principles, strictly in the following format: ``[[rating]]'', e.g. ``[[5]]''.

[Question] \textbf{[QUESTION]}
[Reference answer begins] \textbf{[REFERENCE ANSWER]} [Reference answer ends]

Below are several assistants' answers and their ratings:
[Assistant's answer begins] \textbf{[EXAMPLE ANSWER 1]} [Assistant's answer ends]
Rating: [[\textbf{[RATING 1]}]]
[Assistant's answer begins] \textbf{[EXAMPLE ANSWER 2]} [Assistant's answer ends]
Rating: [[\textbf{[RATING 2]}]]
[Assistant's answer begins] \textbf{[EXAMPLE ANSWER 3]} [Assistant's answer ends]
Rating: [[\textbf{[RATING 3]}]]

Please rate the following assistant answers based on the scoring principles and examples above:
[Assistant's answer begins] \textbf{[RESPONSE]} [Assistant's answer ends]
Rating: \\
\bottomrule
\end{tabular}
\caption{LLM Judge evaluation prompts used to calculate query-level Answer Correctness, adapted to the specific constraints of the LongBench-Cite sub-tasks.}
\label{tab:prompt_correctness}
\end{table}

\section{Methodological Details: Volume Control \& Constraints}
\label{app:methodology}

\subsection{Structural Constraints on Volume Stratification}
The granularity setting $k$ imposes hard lower bounds on which volume ranges can contain statements.
\begin{itemize}[leftmargin=*,noitemsep,topsep=2pt]
    \item \textbf{Absence of Coarse Granularity in Low-Volume Ranges:} At $k{=}8$, a statement citing a single chunk accesses 8 sentences. Therefore, it is mathematically impossible for statements generated at $k{=}8, 16, 32$ to appear in the $2 \le V \le 3$ or $4 \le V \le 7$ volume ranges.
    \item \textbf{Absence of Fine Granularity in High-Volume Ranges:} While theoretically possible, fine-grained settings ($k{=}1$--2) rarely appear in high-volume ranges ($32 \le V \le 63$) because reaching these volumes requires the model to cite a prohibitive number of distinct atomic units (e.g., 32+ separate sentences), which is empirically rare.
\end{itemize}

\subsection{Statistical Validation of Volume Control}
\begin{figure}[t]
    \centering
    \begin{subfigure}{0.48\textwidth}
        \includegraphics[width=\textwidth]{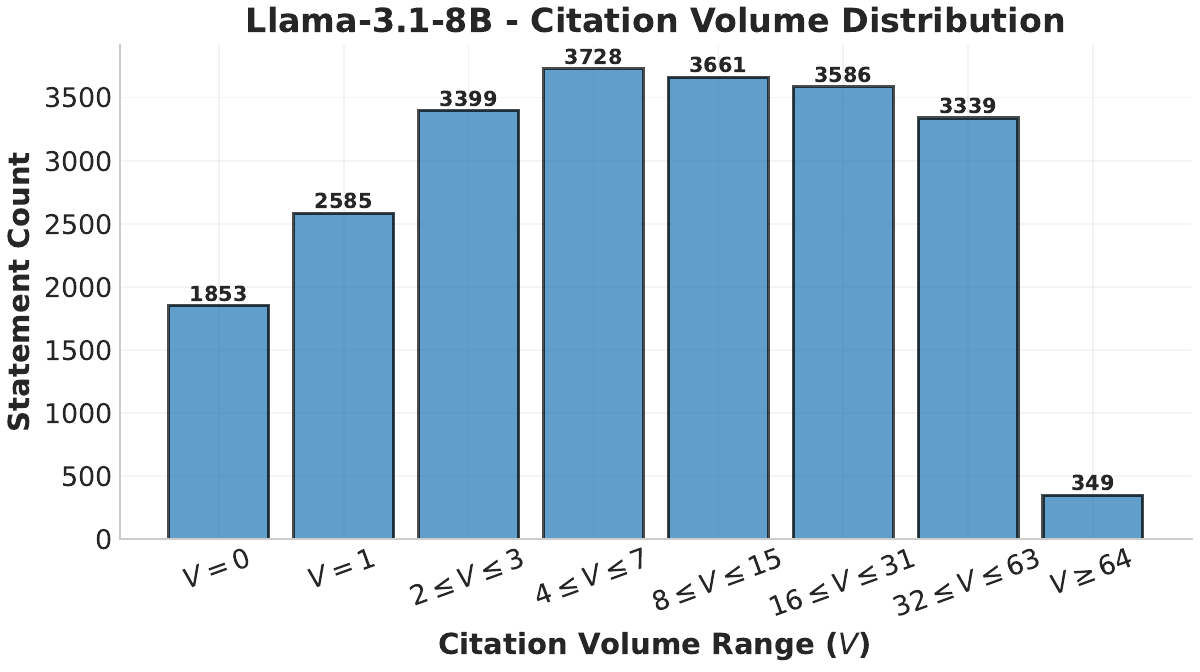}
        \caption{Llama-8B}
    \end{subfigure}
    \hfill
    \begin{subfigure}{0.48\textwidth}
        \includegraphics[width=\textwidth]{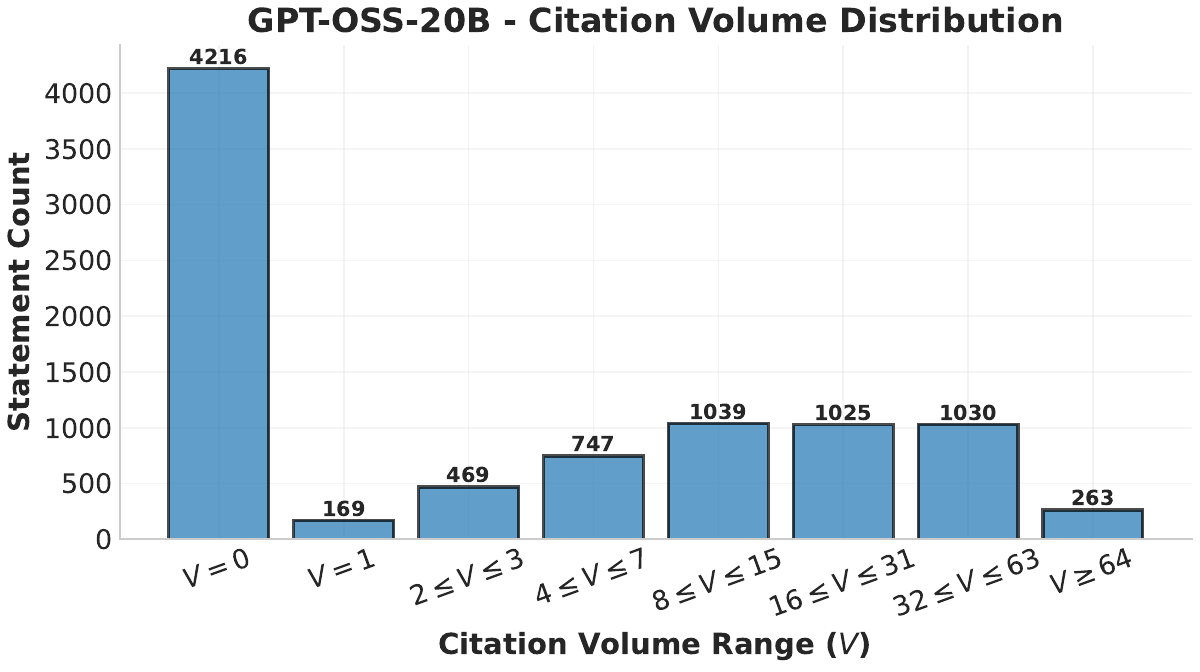}
        \caption{GPT-20B}
    \end{subfigure}
    
    \vspace{1em} 
    
    \begin{subfigure}{0.48\textwidth}
        \includegraphics[width=\textwidth]{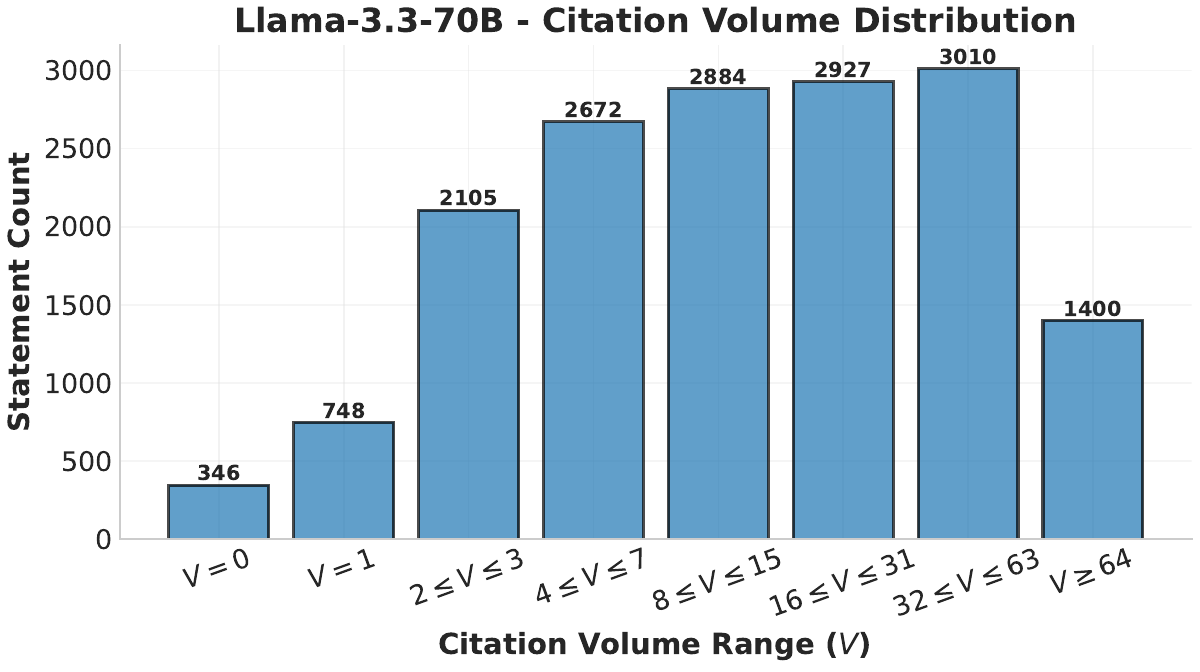}
        \caption{Llama-70B}
    \end{subfigure}
    \hfill
    \begin{subfigure}{0.48\textwidth}
        \includegraphics[width=\textwidth]{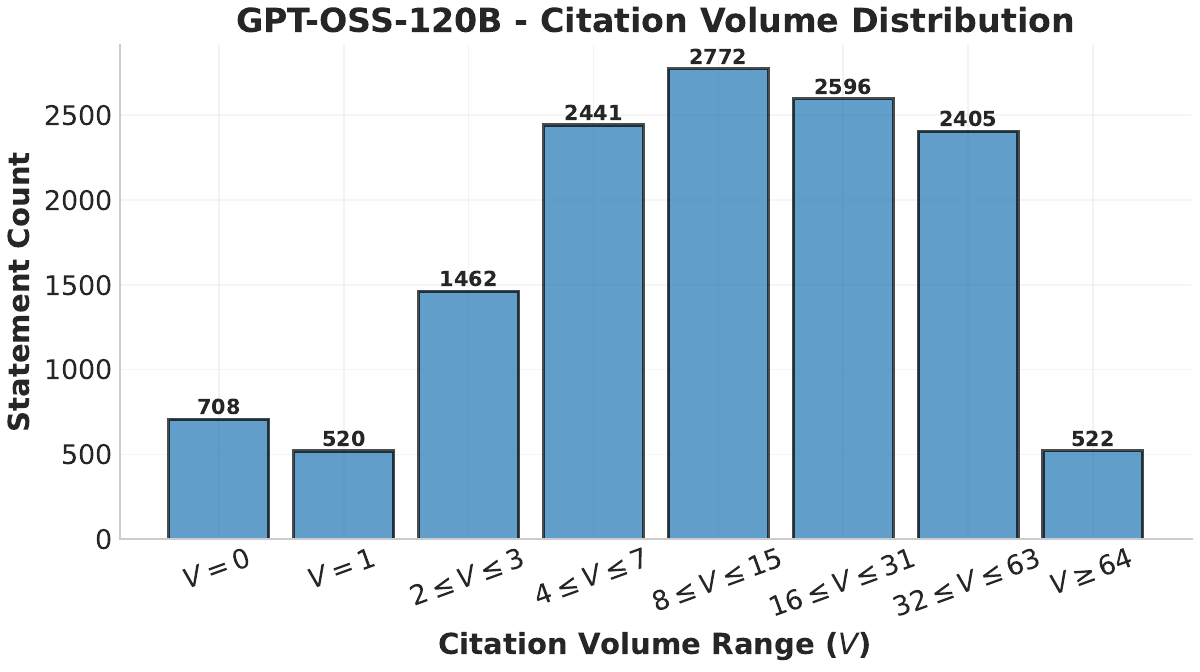}
        \caption{GPT-120B}
    \end{subfigure}
    
    \caption{\textbf{Citation Volume Distributions.} Histograms show statement counts per volume range for each granularity setting. Missing bars reflect the structural constraints described in \S\ref{app:methodology}. Comparisons in the main text are made strictly within vertical slices (fixed volume ranges).}
    \label{fig:app_distribution}
\end{figure}

To ensure our stratification effectively isolates granularity from citation volume, we verify within-range statistics. Taking Llama-70B ($16 \le V \le 31$) as a representative example:
\begin{itemize}[leftmargin=*,noitemsep,topsep=2pt]
    \item \textbf{$k{=}2$}: Mean volume = 19.1 sentences (between-dataset SD = 0.3)
    \item \textbf{$k{=}4$}: Mean volume = 19.0 sentences (between-dataset SD = 0.9)
    \item \textbf{$k{=}8$}: Mean volume = 18.2 sentences (between-dataset SD = 0.4)
\end{itemize}
The tight alignment in mean volumes (19.1 vs 18.2) confirms that statements across different $k$ values cite nearly identical amounts of text within each range. Figure~\ref{fig:app_distribution} shows the full distribution of citation volumes per granularity setting for each model.

\paragraph{Sample Size Variation.}
We note that sample sizes ($N$) vary across settings (e.g., $N{=}225$ for $k{=}2$ vs. $N{=}1141$ for $k{=}8$ in the example above). This imbalance reflects the natural frequency of generation strategies: it is mechanically easier for models to generate a valid $k{=}8$ citation in this volume range than to stitch together 8 separate $k{=}2$ citations. This frequency difference does not invalidate the quality comparison, as we evaluate the \textit{effectiveness} of the attribution when it occurs, not the likelihood of it occurring.

\section{Extended Performance \& Scale Analysis}
\label{app:scale_effects}

This section provides the complete performance data underpinning \S\ref{sec:main_finding} and \S\ref{sec:scale}.

\subsection{Comprehensive Performance Curves}
\begin{figure}[ht]
    \centering
    \begin{subfigure}{0.48\textwidth}
        \includegraphics[width=\textwidth]{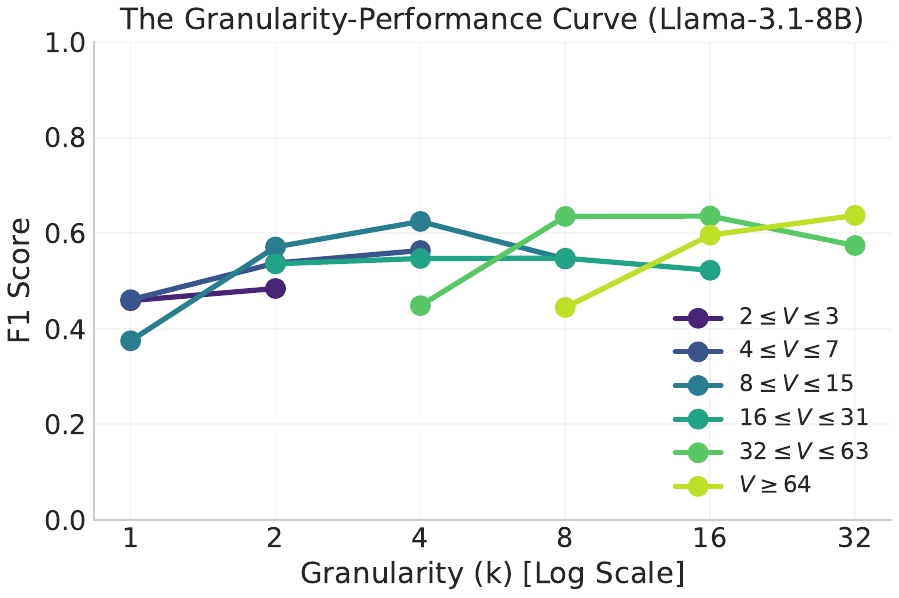}
        \caption{Llama-8B}
    \end{subfigure}
    \hfill
    \begin{subfigure}{0.48\textwidth}
        \includegraphics[width=\textwidth]{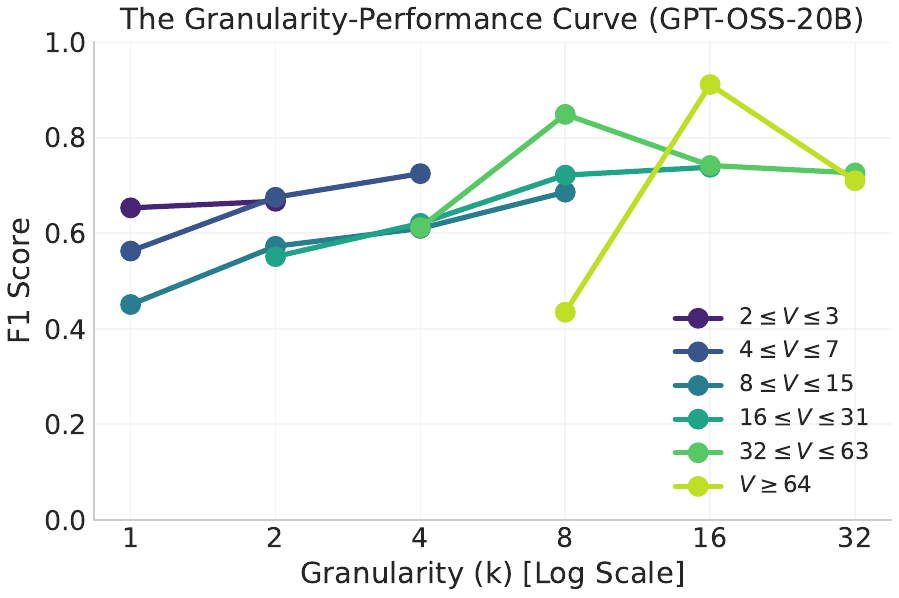}
        \caption{GPT-20B}
    \end{subfigure}

    \vspace{1em} 

    \begin{subfigure}{0.48\textwidth}
        \includegraphics[width=\textwidth]{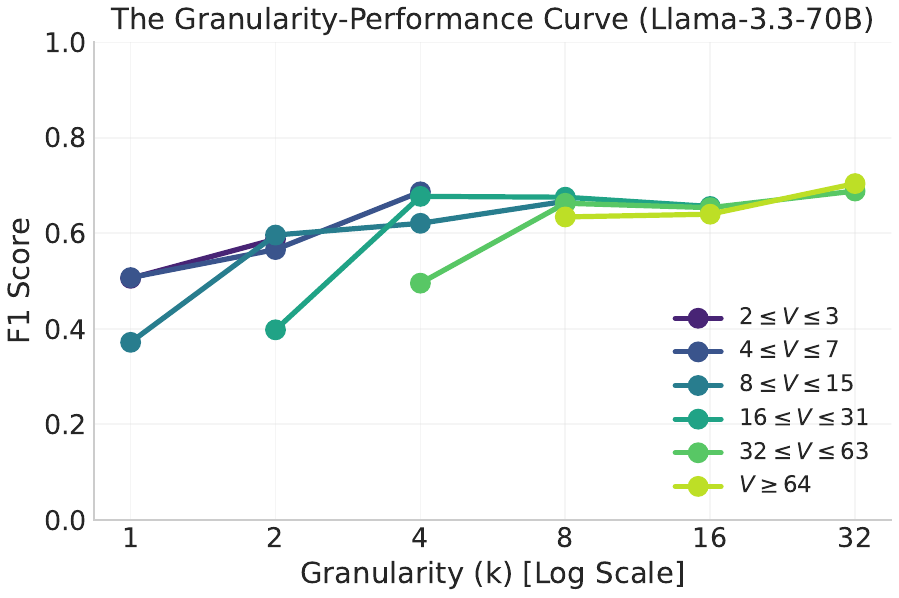}
        \caption{Llama-70B}
    \end{subfigure}
    \hfill
    \begin{subfigure}{0.48\textwidth}
        \includegraphics[width=\textwidth]{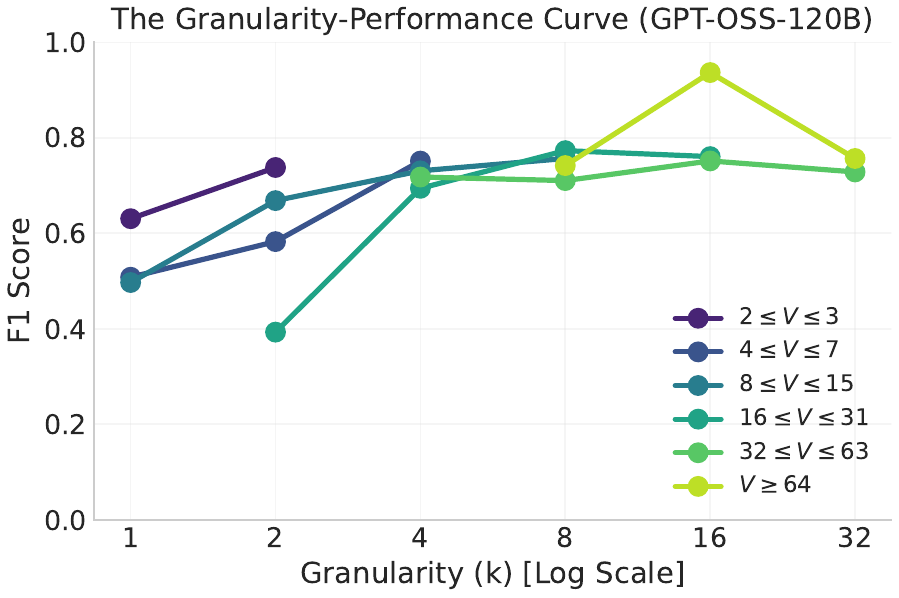}
        \caption{GPT-120B}
    \end{subfigure}

    \caption{\textbf{Comprehensive Granularity-Performance Curves.} Each line represents a distinct citation volume range. Across all four model scales, performance is consistently lowest at fine granularity and peaks at intermediate settings before declining or plateauing at coarse settings.}
    \label{fig:app_phenomenon_others}
\end{figure}

Figure~\ref{fig:main_phenomenon} in the main text highlights the performance curves for three representative volume ranges to demonstrate the intermediate-peaking phenomenon. Figure~\ref{fig:app_phenomenon_others} expands this analysis, presenting the complete set of curves across all valid volume ranges ($V$) for all four models. The universal pattern of performance peaking at intermediate granularities holds across almost all configurations, confirming that the penalty of fine-grained fragmentation is a robust, scale-independent phenomenon.

\subsection{Scale Sensitivity Data}
Table~\ref{tab:scale_sensitivity_complete} provides the comprehensive breakdown of relative and absolute F1 gains across all four volume ranges and all four models.

\begin{table}[ht]
\centering
\small
\begin{tabular}{@{}llcccccc@{}}
\toprule
\multirow{2}{*}{\textbf{Volume ($V$)}} & \multirow{2}{*}{\textbf{Model}} & \multicolumn{2}{c}{\textbf{Fine Baseline}} & \multicolumn{2}{c}{\textbf{Optimal}} & \multicolumn{2}{c}{\textbf{Gains}} \\
\cmidrule(lr){3-4} \cmidrule(lr){5-6} \cmidrule(l){7-8}
& & \textbf{$k$} & \textbf{F1} & \textbf{$k$} & \textbf{F1} & \textbf{Absolute} & \textbf{Relative} \\
\midrule
\multirow{4}{*}{$4 \le V \le 7$} 
& Llama-8B & 1 & 0.461 & 4 & 0.564 & +0.103 & +22.4\% \\
& GPT-20B & 1 & 0.563 & 4 & 0.725 & +0.162 & +28.8\% \\
& Llama-70B & 1 & 0.507 & 4 & 0.687 & +0.180 & +35.4\% \\
& GPT-120B & 1 & 0.508 & 4 & 0.751 & +0.243 & +47.9\% \\
\midrule
\multirow{4}{*}{$8 \le V \le 15$} 
& Llama-8B & 1 & 0.375 & 4 & 0.625 & +0.250 & +66.7\% \\
& GPT-20B & 1 & 0.451 & 8 & 0.686 & +0.235 & +52.2\% \\
& Llama-70B & 1 & 0.372 & 8 & 0.667 & +0.295 & +79.5\% \\
& GPT-120B & 1 & 0.497 & 8 & 0.756 & +0.260 & +52.3\% \\
\midrule
\multirow{4}{*}{$16 \le V \le 31$} 
& Llama-8B & 2 & 0.536 & 8 & 0.548 & +0.012 & +2.2\% \\
& GPT-20B & 2 & 0.551 & 16 & 0.738 & +0.187 & +34.0\% \\
& Llama-70B & 2 & 0.398 & 4 & 0.677 & +0.279 & +70.2\% \\
& GPT-120B & 2 & 0.393 & 8 & 0.773 & +0.380 & +96.7\% \\
\midrule
\multirow{4}{*}{$32 \le V \le 63$\textsuperscript{†}} 
& Llama-8B & 4 & 0.448 & 16 & 0.636 & +0.188 & +41.8\% \\
& GPT-20B & 4 & 0.612 & 8 & 0.849 & +0.237 & +38.7\% \\
& Llama-70B & 4 & 0.495 & 32 & 0.688 & +0.193 & +38.9\% \\
& GPT-120B & 4 & 0.718 & 16 & 0.751 & +0.034 & +4.7\% \\
\bottomrule
\end{tabular}
\caption{\textbf{Complete Scale-Dependent Granularity Effects.} \textsuperscript{†}Uses $k{=}4$ baseline as fine-grained citations ($k{=}1,2$) rarely reach this volume (see Appendix~\ref{app:methodology}).}
\label{tab:scale_sensitivity_complete}
\end{table}

\section{Mechanistic Analysis: Precision vs. Recall}
\label{app:precision_recall}

This section details the Precision-Recall trade-off described in \S\ref{sec:mechanism}.

\subsection{Detailed Pattern Analysis}
Table~\ref{tab:mechanism_comprehensive} presents a comprehensive breakdown of the mechanistic patterns for all models across all volume ranges. We examine two key properties:
\begin{enumerate}
    \item \textbf{Precision Monotonicity:} Does precision strictly increase (or stay flat) as granularity becomes coarser? This indicates \textit{Boundary Tolerance}.
    \item \textbf{Recall Peak \& Decline:} At what granularity ($k$) does recall peak, and does it decline at the coarsest settings? This indicates \textit{Signal Dilution}.
\end{enumerate}

\paragraph{Universality.} Precision monotonicity holds in 94\% (15/16) of cases. The recall ``peak-and-decline'' pattern also holds in 94\% (15/16) of cases, confirming that signal dilution is a robust phenomenon. We exclude extreme volume ranges ($2 \le V \le 3$ and $V \ge 64$) from this specific analysis, as they lack sufficient granularity settings or sample sizes to construct reliable trend curves.

\paragraph{Volume-Dependent Shift.} As shown in the ``Recall Pattern'' column of Table~\ref{tab:mechanism_comprehensive}, the granularity required to maximize recall shifts rightward (to coarser settings) as citation volume increases. For example, Llama-70B's peak moves from $k{=}1$ in low-volume settings to $k{=}8$ in high-volume settings. This confirms that models require coarser chunks to maintain coverage when processing larger amounts of evidence.

\begin{table}[ht]
\centering
\small
\begin{tabular}{@{}llcc@{}}
\toprule
\textbf{Model} & \textbf{Volume ($V$)} & \textbf{Precision Monotonicity} & \textbf{Recall Pattern (Peak $k$)} \\
\midrule
\multirow{4}{*}{Llama-3.1-8B} 
 & $4 \le V \le 7$   & \checkmark & Peak $k{=}1$ $\to$ Decline \\
 & $8 \le V \le 15$  & \checkmark & Peak $k{=}2$ $\to$ Decline \\
 & $16 \le V \le 31$ & \checkmark & Peak $k{=}2$ $\to$ Decline \\
 & $32 \le V \le 63$ & \checkmark & Peak $k{=}8$ $\to$ Decline \\
\midrule
\multirow{4}{*}{GPT-OSS-20B} 
 & $4 \le V \le 7$   & \checkmark & Peak $k{=}1$ $\to$ Decline \\
 & $8 \le V \le 15$  & \checkmark & Peak $k{=}1$ $\to$ Decline \\
 & $16 \le V \le 31$ & \checkmark & Peak $k{=}2$ $\to$ Decline \\
 & $32 \le V \le 63$ & $\times$ (Fluctuates) & Peak $k{=}8$ $\to$ Decline \\
\midrule
\multirow{4}{*}{Llama-3.3-70B} 
 & $4 \le V \le 7$   & \checkmark & Peak $k{=}1$ $\to$ Decline \\
 & $8 \le V \le 15$  & \checkmark & Peak $k{=}2$ $\to$ Decline \\
 & $16 \le V \le 31$ & \checkmark & Peak $k{=}4$ $\to$ Decline \\
 & $32 \le V \le 63$ & \checkmark & Peak $k{=}8$ $\to$ Decline \\
\midrule
\multirow{4}{*}{GPT-OSS-120B} 
 & $4 \le V \le 7$   & \checkmark & Monotonic Increase (Peak $k{=}4$) \\
 & $8 \le V \le 15$  & \checkmark & Peak $k{=}2$ $\to$ Decline \\
 & $16 \le V \le 31$ & \checkmark & Peak $k{=}8$ $\to$ Decline \\
 & $32 \le V \le 63$ & \checkmark & Peak $k{=}4$ $\to$ Decline \\
\bottomrule
\end{tabular}
\caption{\textbf{Comprehensive Mechanistic Analysis.} We observe distinct patterns: (1) Precision is monotonic in 15/16 cases. (2) Recall peaks shift to coarser granularities (higher $k$) as citation volume increases (moving down the rows for each model).}
\label{tab:mechanism_comprehensive}
\end{table}

\subsection{Full Metric Decomposition}
\begin{figure}[p]
    \centering
    \includegraphics[width=0.95\textwidth]{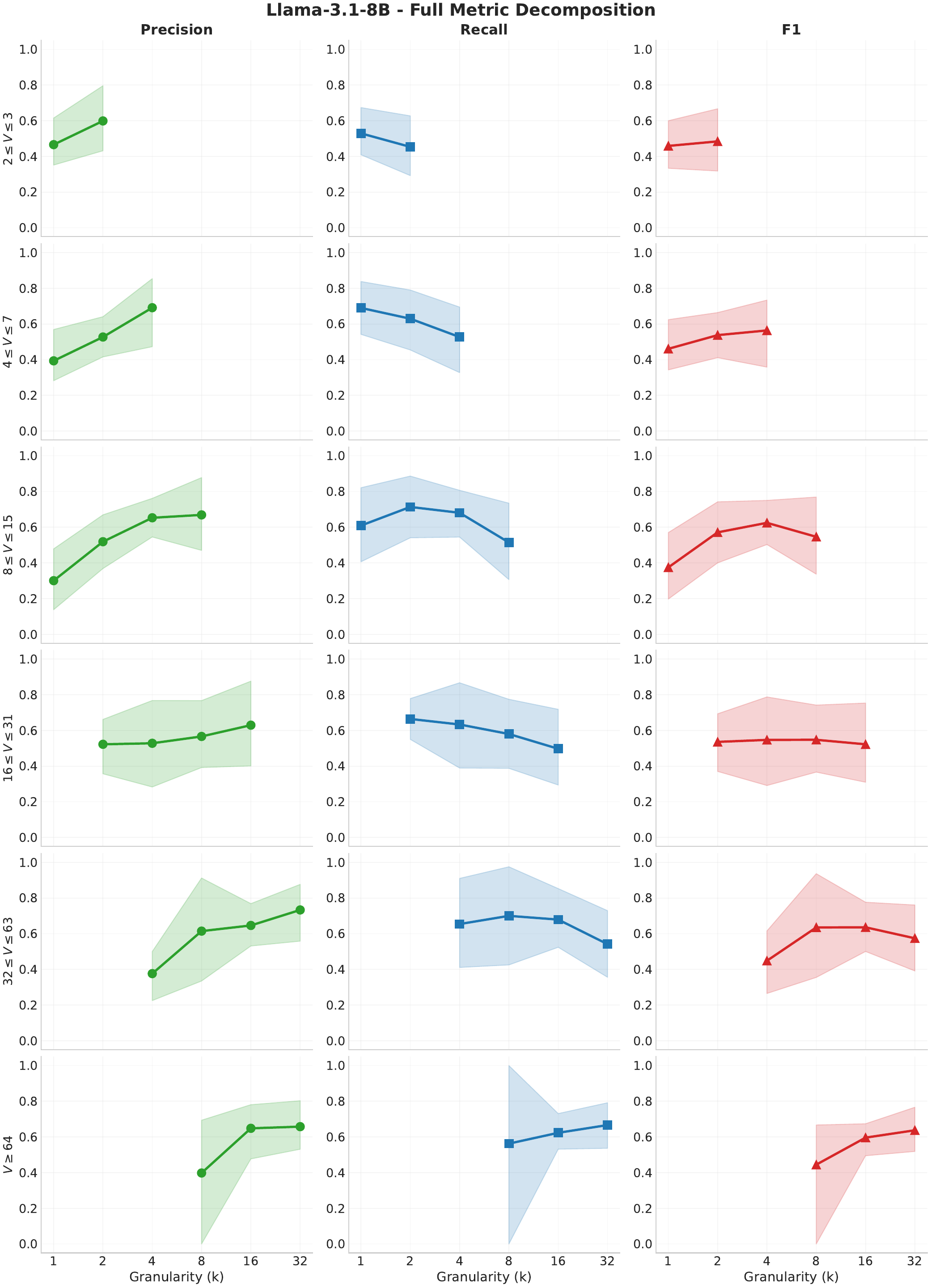}
    \caption{\textbf{Llama-8B Decomposition.} Precision (green), Recall (blue), F1 (red).}
    \label{fig:grid_llama8b}
\end{figure}

\begin{figure}[p]
    \centering
    \includegraphics[width=0.95\textwidth]{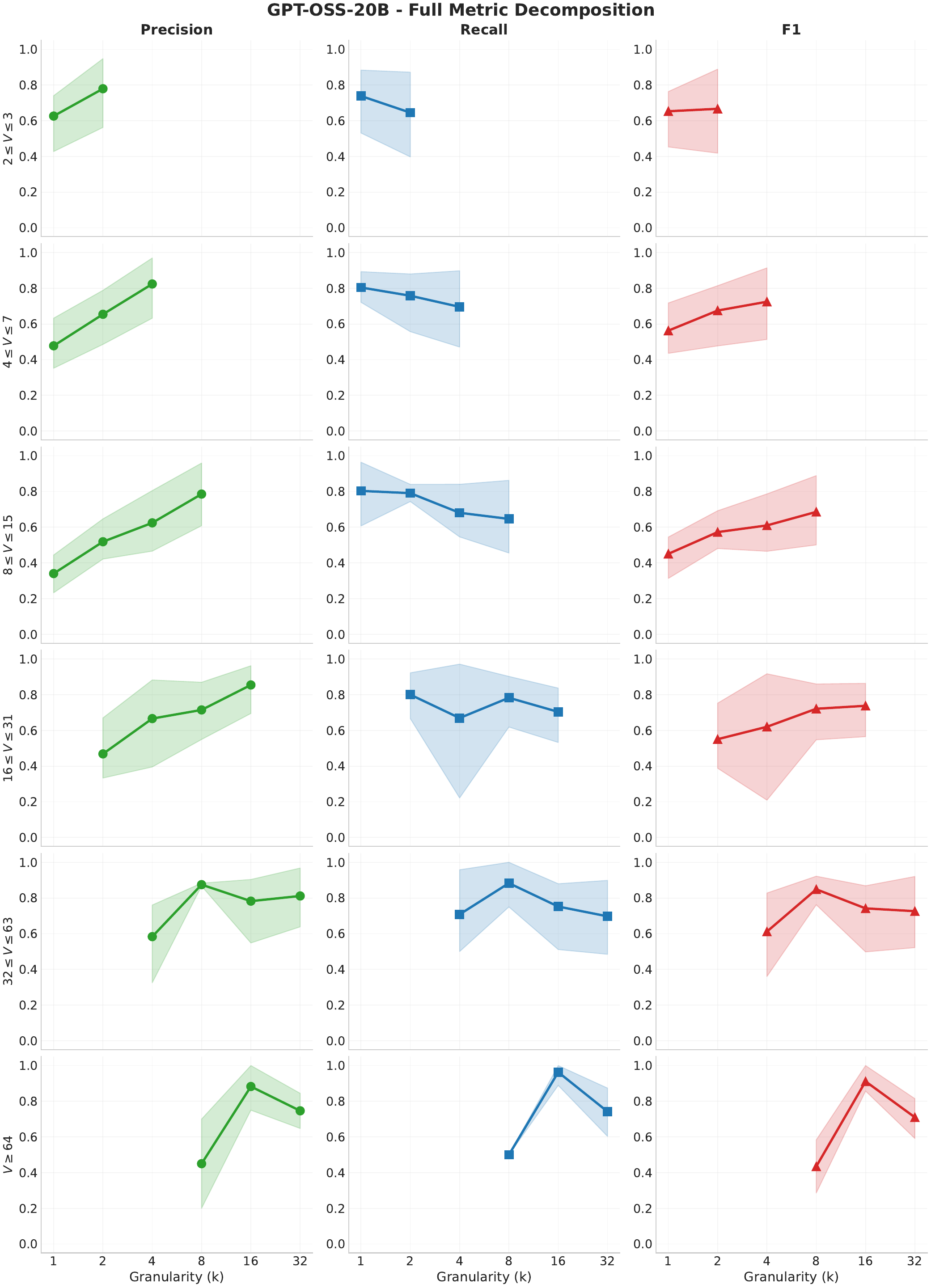}
    \caption{\textbf{GPT-20B Decomposition.} Precision (green), Recall (blue), F1 (red).}
    \label{fig:grid_gpt20b}
\end{figure}

\begin{figure}[p]
    \centering
    \includegraphics[width=0.95\textwidth]{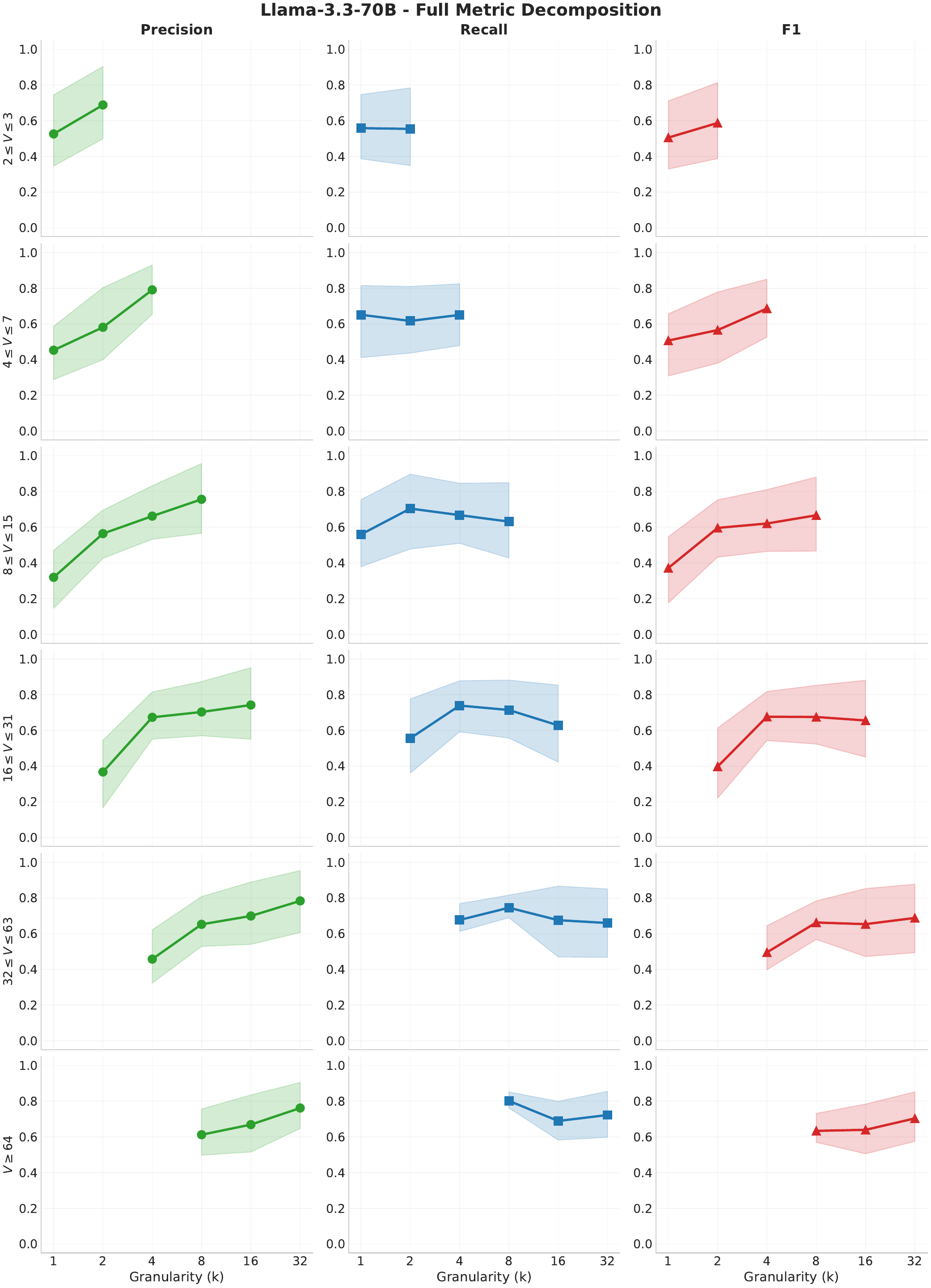}
    \caption{\textbf{Llama-70B Decomposition.} Precision (green), Recall (blue), F1 (red).}
    \label{fig:grid_llama70b}
\end{figure}

\begin{figure}[p]
    \centering
    \includegraphics[width=0.95\textwidth]{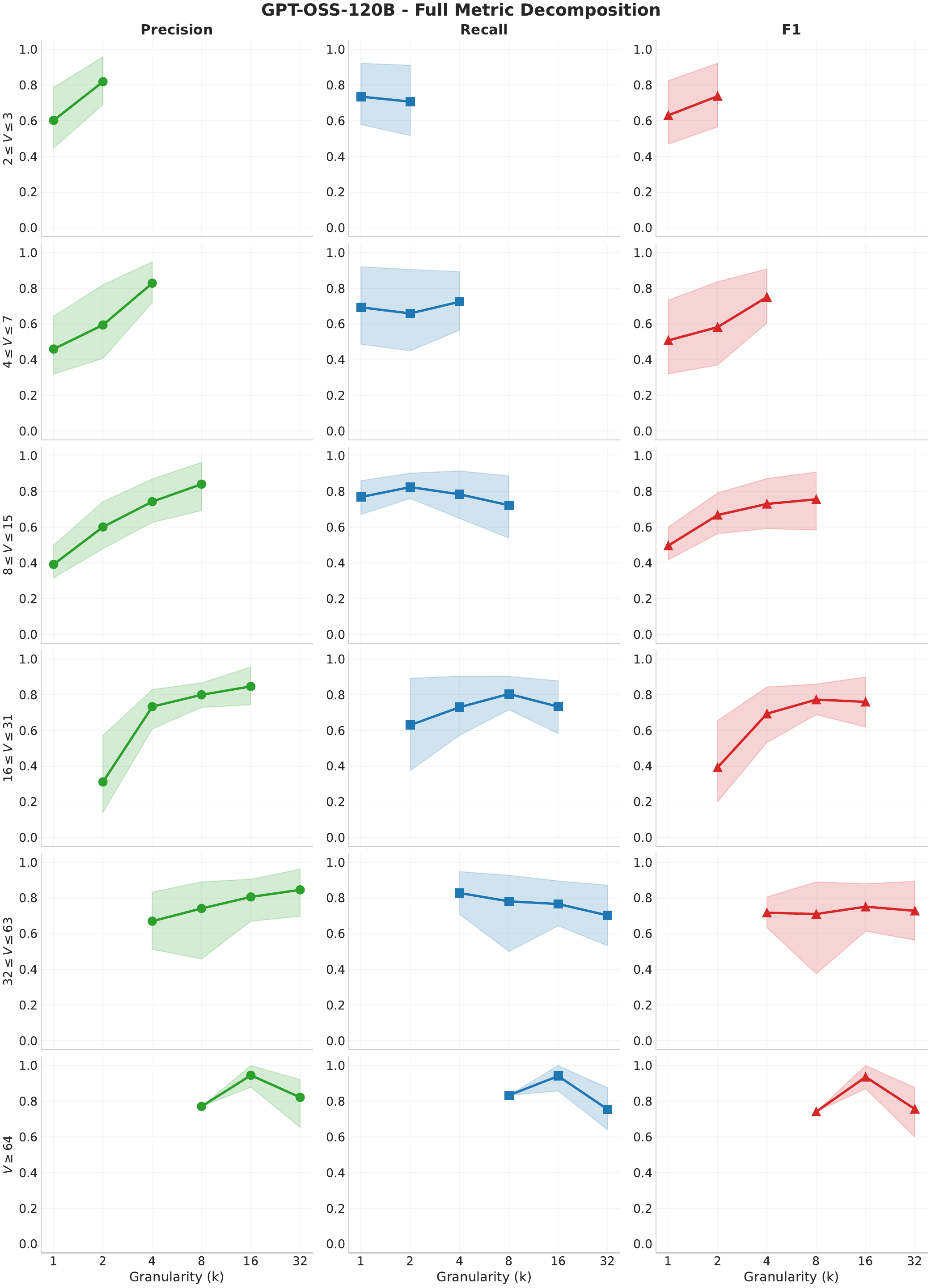}
    \caption{\textbf{GPT-120B Decomposition.} Precision (green), Recall (blue), F1 (red).}
    \label{fig:grid_gpt120b}
\end{figure}

Figures~\ref{fig:grid_llama8b}--\ref{fig:grid_gpt120b} visualize these trade-offs.

\section{Task-Dependent Analysis}
\label{app:tasks}

This section expands the analysis in \S\ref{sec:generalization} to all datasets and models.

Table~\ref{tab:dataset_comprehensive} presents the citation-optimal granularity and relative gains for all model-dataset pairs in the medium-volume range ($16 \le V \le 31$). Consistent with the main text, we observe that optimal granularity shifts based on task requirements:
\begin{itemize}[leftmargin=*,noitemsep,topsep=2pt]
    \item \textbf{Summarization (GovReport):} Favors coarse granularity ($k{=}16$) for three of the four models, allowing models to synthesize information from larger contiguous blocks.
    \item \textbf{Multi-hop QA (HotpotQA):} Consistently favors intermediate granularity ($k{=}4$), balancing precise evidence isolation with sufficient context.
    \item \textbf{Factoid \& Chat (MultiField, LB-Chat):} Vary by model, but reject fine-grained ($k{=}2$) settings in all but one cell, in favor of broader context ($k{=}4, 8,$ or $16$).
\end{itemize}

\begin{table}[ht]
\centering
\small
\setlength{\tabcolsep}{4pt} 
\begin{tabular}{@{}lllccccccccc@{}}
\toprule
\multirow{2}{*}{\textbf{Model}} & \multirow{2}{*}{\textbf{Volume ($V$)}} & \multirow{2}{*}{\textbf{Dataset}} & \multicolumn{6}{c}{\textbf{F1 at Granularity ($k$)}} & \multirow{2}{*}{\textbf{Opt $k$}} & \multirow{2}{*}{\textbf{Opt F1}} & \multirow{2}{*}{\textbf{Gain}} \\
\cmidrule(lr){4-9}
& & & \textbf{1} & \textbf{2} & \textbf{4} & \textbf{8} & \textbf{16} & \textbf{32} & & & \\
\midrule
\multirow{4}{*}{Llama-70B} 
& \multirow{4}{*}{$16 \le V \le 31$} 
  & GovReport  & --- & 0.653 & 0.826 & 0.879 & \textbf{0.901} & --- & 16 & 0.901 & +37.9\% \\
& & HotpotQA   & --- & 0.315 & \textbf{0.602} & 0.565 & 0.459 & --- &  4 & 0.602 & +91.1\% \\
& & LB-Chat    & --- & 0.129 & \textbf{0.485} & 0.482 & 0.443 & --- &  4 & 0.485 & +276\% \\
& & MultiField & --- & 0.494 & 0.795 & 0.775 & \textbf{0.821} & --- & 16 & 0.821 & +66.4\% \\
\midrule
\multirow{4}{*}{GPT-120B} 
& \multirow{4}{*}{$16 \le V \le 31$} 
  & GovReport  & --- & 0.725 & 0.881 & 0.876 & \textbf{0.921} & --- & 16 & 0.921 & +27.1\% \\
& & HotpotQA   & --- & 0.234 & \textbf{0.697} & 0.674 & 0.596 & --- &  4 & 0.697 & +198\% \\
& & LB-Chat    & --- & 0.167 & 0.466 & \textbf{0.730} & 0.689 & --- &  8 & 0.730 & +338\% \\
& & MultiField & --- & 0.447 & 0.732 & 0.811 & \textbf{0.836} & --- & 16 & 0.836 & +87.2\% \\
\midrule
\multirow{4}{*}{Llama-8B} 
& \multirow{4}{*}{$16 \le V \le 31$} 
  & GovReport  & --- & 0.744 & \textbf{0.870} & 0.756 & 0.772 & --- &  4 & 0.870 & +17.0\% \\
& & HotpotQA   & --- & 0.543 & \textbf{0.584} & 0.451 & 0.355 & --- &  4 & 0.584 & +7.6\% \\
& & LB-Chat    & --- & \textbf{0.545} & 0.194 & 0.282 & 0.265 & --- &  2 & 0.545 & +0.0\% \\
& & MultiField & --- & 0.312 & 0.541 & \textbf{0.702} & 0.698 & --- &  8 & 0.702 & +125\% \\
\midrule
\multirow{6}{*}{Llama-70B} 
& \multirow{2}{*}{$4 \le V \le 7$}
  & GovReport  & 0.652 & 0.807 & \textbf{0.865} & --- & --- & --- &  4 & 0.865 & +32.6\% \\
& & HotpotQA   & \textbf{0.524} & 0.380 & 0.516 & --- & --- & --- &  1 & 0.524 & +0.0\% \\
\cmidrule(lr){2-12}
& \multirow{2}{*}{$8 \le V \le 15$}
  & GovReport  & 0.595 & 0.730 & 0.845 & \textbf{0.894} & --- & --- &  8 & 0.894 & +50.4\% \\
& & HotpotQA   & 0.389 & \textbf{0.561} & 0.510 & 0.482 & --- & --- &  2 & 0.561 & +44.0\% \\
\cmidrule(lr){2-12}
& \multirow{2}{*}{$32 \le V \le 63$}
  & GovReport  & --- & --- & 0.717 & 0.810 & 0.875 & \textbf{0.898} & 32 & 0.898 & +25.2\% \\
& & HotpotQA   & --- & --- & 0.377 & \textbf{0.608} & 0.586 & 0.467 &  8 & 0.608 & +61.2\% \\
\bottomrule
\end{tabular}
\caption{\textbf{Comprehensive dataset performance analysis.} Includes $k{=}1$ for relevant volume ranges. Gains are calculated as $(\text{Opt F1} - \text{Finest F1}) / \text{Finest F1}$, where ``Finest F1'' corresponds to the smallest granularity available in that range (e.g., $k{=}1$ for $4 \le V \le 7$, $k{=}2$ for $16 \le V \le 31$, $k{=}4$ for $32 \le V \le 63$). This metric quantifies the improvement of optimal settings over the finest-grained constraint possible for that volume.}
\label{tab:dataset_comprehensive}
\end{table}

\section{Extended Correctness Analysis}
\label{app:correctness}

This section expands on the asymmetric sensitivity findings in \S\ref{sec:correctness}, examining the trade-off between attribution quality and answer correctness across all models.

\subsection{Alignment and Trade-offs}
Table~\ref{tab:correctness_complete} compares the correctness scores at the \textbf{Citation-Optimal} granularity (bold) versus the absolute maximum correctness achievable in that range.

We observe that while exact alignment of optima is rare (25\% of cases), the cost of misalignment is generally low. For capable models like Llama-70B, the correctness score at the citation-optimal $k$ is within 0.9\% of the absolute maximum in aggregate. Localized trade-offs appear primarily in the medium-volume range for Llama-8B, GPT-20B, and GPT-120B, where optimizing for citation quality ($k{\geq}8$) can incur a correctness penalty of 14--22\% compared to finer-grained settings. However, these are exceptions; for the vast majority of configurations, citation quality can be maximized with negligible impact on accuracy. We note that correctness is scored once per query over the complete answer, whereas volume ranges partition individual statements; a query therefore contributes its score to several ranges at once. The aggregate comparison in Table~\ref{tab:asymmetry}a, which requires no such attribution, is the basis for our claim, and we report the breakdown below for completeness.

\begin{table}[ht]
\centering
\small
\setlength{\tabcolsep}{4pt}
\begin{tabular}{@{}llccccccc@{}}
\toprule
\multirow{2}{*}{\textbf{Model}} & \multirow{2}{*}{\textbf{Volume ($V$)}} & \multicolumn{6}{c}{\textbf{Answer Correctness at Granularity}} & \multirow{2}{*}{\textbf{Cit. Opt $k$}} \\
\cmidrule(lr){3-8}
& & \textbf{$k{=}1$} & \textbf{$k{=}2$} & \textbf{$k{=}4$} & \textbf{$k{=}8$} & \textbf{$k{=}16$} & \textbf{$k{=}32$} & \\
\midrule
\multirow{4}{*}{Llama-8B} 
& $4 \le V \le 7$   & 0.660 & 0.665 & \textbf{0.606} & ---   & ---   & ---   & 4 \\
& $8 \le V \le 15$  & 0.658 & 0.662 & \textbf{0.603} & 0.647 & ---   & ---   & 4 \\
& $16 \le V \le 31$ & ---   & 0.817 & 0.648 & \textbf{0.634} & 0.600 & ---   & 8 \\
& $32 \le V \le 63$ & ---   & ---   & 0.574 & 0.709 & \textbf{0.649} & 0.621 & 16 \\
\midrule
\multirow{4}{*}{Llama-70B} 
& $4 \le V \le 7$   & 0.689 & 0.734 & \textbf{0.730} & ---   & ---   & ---   & 4 \\
& $8 \le V \le 15$  & 0.677 & 0.730 & 0.716 & \textbf{0.739} & ---   & ---   & 8 \\
& $16 \le V \le 31$ & ---   & 0.676 & \textbf{0.692} & 0.728 & 0.726 & ---   & 4 \\
& $32 \le V \le 63$ & ---   & ---   & 0.711 & 0.717 & 0.733 & \textbf{0.716} & 32 \\
\midrule
\multirow{4}{*}{GPT-120B} 
& $4 \le V \le 7$   & 0.546 & 0.587 & \textbf{0.629} & ---   & ---   & ---   & 4 \\
& $8 \le V \le 15$  & 0.628 & 0.608 & 0.577 & \textbf{0.603} & ---   & ---   & 8 \\
& $16 \le V \le 31$ & ---   & 0.699 & 0.610 & \textbf{0.588} & 0.669 & ---   & 8 \\
& $32 \le V \le 63$ & ---   & ---   & 0.673 & 0.570 & \textbf{0.660} & 0.656 & 16 \\
\midrule
\multirow{4}{*}{GPT-20B} 
& $4 \le V \le 7$   & 0.556 & 0.585 & \textbf{0.644} & ---   & ---   & ---   & 4 \\
& $8 \le V \le 15$  & 0.523 & 0.583 & 0.634 & \textbf{0.537} & ---   & ---   & 8 \\
& $16 \le V \le 31$ & ---   & 0.669 & 0.632 & 0.620 & \textbf{0.574} & ---   & 16 \\
& $32 \le V \le 63$ & ---   & ---   & 0.631 & \textbf{0.641} & 0.635 & 0.568 & 8 \\
\bottomrule
\end{tabular}
\caption{\textbf{Correctness scores across all settings.} Bold values indicate the correctness score achieved at the \textbf{Citation-Optimal} granularity. Note that structural constraints (see \S\ref{app:methodology}) prevent certain granularities from appearing in specific volume ranges.}
\label{tab:correctness_complete}
\end{table}

\end{document}